\documentclass{article}
\usepackage{iclr2026_conference,times,lila}
\preprint 
\usepackage{graphicx}
\usepackage{float}
\usepackage{array}
\usepackage{longtable}
\usepackage{subcaption}
\usepackage{hyperref}
\usepackage{listings}
\usepackage{comment}
\usepackage{booktabs}
\usepackage{amsmath}
\usepackage{multirow}
\usepackage{epigraph}
\usepackage{flafter}
\usepackage{placeins}
\title{%
The Value of Not Knowing: How Fresh Eyes through Naive LLMs Fuel Creativity
}

\author{David B.\ D'Ambrosio, Alon Albalak, Joel Lehman, Kenneth O.\ Stanley}
\date{October 2026}

\begin{document}

\maketitle

\begin{abstract}
Many major scientific advances come from approaching a problem with a new or novel perspective.  While large language models (LLMs) with internet-scale knowledge seem poised to take such perspectives, their consistent generation remains elusive. Instead, models are drawn to dominant views and preexisting, in-field bias.  To explore the potential of fresh eyes to increase creativity, this paper proposes \emph{naive LLMs}, models that are made intentionally ignorant of a specific topic.  Through deprivation in training or intentional forgetting, naive models are unburdened by existing dogma, allowing them to be free to make new associations and conjectures.  These naive models can then be paired with a stronger \emph{mentor} model that can refine their unique perspectives into concrete proposals. For the 1B and 7B models tested in this paper, naive models consistently produce seed ideas that are more novel and useful than baseline counterparts, and mentor refinement improves the ideas even further.  Thus, contrary to the prevailing tradition of scaling up training data, we demonstrate a case where \emph{less} information is actually more powerful, in particular when the aim is to discover  genuinely outside-the-box solutions.
\end{abstract}

\epigraph{If we approach a task with ignorance, it can remove the barricade of knowledge blocking progress. }%
{---\textup{Rick Rubin}, \textup{\emph{The Creative Act}}~\citep{rubin2023creative}}

\section{Introduction} \label{sec:intro}

With so much emphasis today on scaling laws~\citep{kaplan2020scaling, subramanyam2026scalinglawsrevisitedmodeling} and data acquisition~\citep{villalobos2024position}, the possibility that less data or \emph{knowing less} could ever be advantageous is rarely discussed.  Yet there is at least one intriguing precedent with humans for questioning the assumption that more is always better: in particular, students or newcomers to fields often benefit from the novel perspective possible through \emph{fresh eyes}.  Indeed, throughout the history of science there are many famous examples of newcomers to a field who made breakthrough discoveries because they were unencumbered by typical biases or brought a fresh perspective from a different area of study~\citep{jeppesen2010marginality, azoulay2019funeral, kuhn2012structure}.  

The advantage of fresh eyes is particularly provocative in the context of today's relentless quest for ever more data and scale.  What does it mean for LLMs that it is sometimes helpful for humans to be naive in a particular area, particularly in the pursuit of creative insight~\citep{wiley1998expertise, jeppesen2010marginality}?  While the ingredients for unconventional creative insight may in principle exist within the weights of a large model, overwhelming familiarity with the dominant paradigm may make such connections prohibitively unlikely.  

The issue is not that current models are incapable of \emph{all} creative insight.  
LLMs have produced research contributions across multiple fields~\citep{alon2026remarks,tsoukalas2026alphaproof,guevara2026gluon,openai2026math}.
Rather, the opportunity is that \emph{some} kinds of insights are more likely through fresh eyes, and those are the types that current monolithic models are not suited to discovering.
The possibility of such ``naive insight'' represents a particularly delicate research opportunity because it pushes against the tidal wave of scaling and data that currently propels almost every measure of progress in the field.
After all, the bitter lesson~\citep{sutton2019bitter, kaplan2020scaling, hoffmann2022empirical} is often interpreted to imply that more computation and more data will continue to benefit models.
To gain even a foothold on this opportunity to explore the implications of fresh eyes requires at least a partial abandonment of well-understood best practices and the idea that ``knowing everything'' is always the best approach.

So in what way should the potential of fresh eyes be explored? Prompting a model to approach problems pretending it has no knowledge is an option, but models have difficulty suppressing what they know when instructed to~\citep{li2026simulated}, and even when made to underperform on command they only imperfectly imitate a genuinely less capable model~\citep{vanderweij2024sandbagging}; the information is still there and will ultimately influence the response. Results in this paper further confirm the limitations of such a simplistic prompt-based approach.

As an alternative, the approach that follows draws inspiration from the notion of an outsider or young person entering a new field by introducing the concept of a \emph{naive LLM}, which means an LLM that is unaware of a specific topic while maintaining a scale-appropriate understanding of everything else.  The idea that the model should still be competent outside the area of ignorance is important: we would not expect innovative thinking to emerge from generally diminished cognition. On the contrary, the idea is that an otherwise competent intellect is particularly empowered when unleashed into a new area (to it) with fresh eyes, unlike standard models that have been shown to be drawn to existing distributions~\citep{zhang2026verbalized, kirk2024understanding}.

In this way, naive LLMs differ from other attempts at limited models such as LittleLearner~\citep{li2026littlelearner} and the GPT-1900 model~\citep{hla2026machina} in the sense that naive models still represent a capable individual with modern knowledge and can apply these vast capabilities to an ``unfamiliar'' domain.  By limiting the model’s knowledge selectively rather than broadly, it can draw on a breadth of skills and expertise that can be applied free of preconceptions to its domain of ignorance.

Furthermore, we also explore the potential for the advantage of fresh eyes to be \emph{refined} with the help of a more knowledgeable \emph{mentor}.  While a naive entrant to a field may have interesting seed ideas, refining those ideas into more practical proposals can take some guidance.  The intriguing possibility is that once that guidance is provided, the advantage of coming from a fresh perspective may be made more practical.

Two different methods for eliciting naive behavior from LLMs are explored: training from scratch and forgetting. These approaches are tested across model sizes, model families, and topics.  The compute necessary to train a new model from scratch or even to make it forget a topic means that naive networks here are restricted to smaller 1B and 7B sizes, but the results still provide a strong hint that \emph{less can be more}.  As long as topics are not beyond the cognitive capacity of these small models, a consistent trend emerges suggesting the very unusual result that more ignorant models are genuinely scoring higher in creative measures and separating themselves farther out in embedding space.
Mentors also help refine the utility and feasibility of the naively originated ideas.

The aim of this work is not to be prescriptive, that is, there are many possible definitions of and ways to make a naive LLM, even beyond what is introduced here. Instead the results that follow shine a light on the question of whether it is better to know ``everything,'' and suggest that there are potentially ways to benefit from having less knowledge, especially in the domain of creativity. While only a start, these results matter because they raise the intriguing question of what a frontier model that had never seen the oncology literature might propose about cancer research.

\section{Background} 
This section reviews topics related to success with limited knowledge, including a general scientific view, past work in machine learning on systems with limited knowledge, and attempts to reduce the knowledge of existing models.

\label{sec:background}

\subsection{The Downsides (and Upsides) of Knowledge in Science}

The Wright Brothers ran a bicycle repair shop and lacked any formal engineering training, and yet they managed to create the world's first true airplane~\citep{jakab1990visions}.  Science is filled with such examples of approaching problems with ``fresh eyes'' and proposing new solutions unbounded by the typical biases and beliefs of the field.  However, these are not just black swan anecdotes; research has shown that it is exactly these new perspectives that lead to breakthroughs.  \citet{wiley1998expertise} demonstrates that strong existing domain knowledge can constrain experts to ``an area of the search space in which the solution does not reside'' when confronted with problems related to that domain.  Additionally, an analysis of a scientific bounty program, InnoCentive, showed an unintuitively \emph{positive} correlation between a solver's success and the distance between the field of the question and the field of the scientist answering it, on problems that had resisted traditional approaches~\citep{jeppesen2010marginality}.  Furthermore, when established scientists pass away there is a documented surge in new and novel approaches in their field~\citep{azoulay2019funeral}. 

These new and novel approaches typically come from outside the establishment; either from other fields or from young scientists entering the field. Indeed, inventive and scientific disruptiveness declines with age~\citep{kaltenberg2023invention, cui2026aging}.   Younger researchers are also more likely to depart from existing approaches~\citep{packalen2019age}.  Still, peak research output remains later in life and its age has been trending higher in recent decades~\citep{jones2010age}, which is attributed to (human) training time.  So competency and creativity move in opposite directions, while the growing corpus of knowledge means sacrificing a researcher's most creative and open years to additional study.  This paper decouples these seemingly opposing forces into the naive and mentor roles.

\citet{uzzi2013atypical} studied millions of papers in an attempt to determine what makes research successful, and found that the highest-impact work pairs a conventional base with a few unusual, novel combinations.  In fact, exposure to a persistent minority viewpoint can stimulate divergent thinking in a majority and lead them to correct solutions the minority never even offered~\citep{nemeth1986differential, nemeth1983creative}, and diverse groups can sometimes outperform skilled ones~\citep{hong2004groups}.  These examples all show that ideation unencumbered with conventional dogma can sometimes unlock productive new research directions.

\subsection{Machine Learning with Limited Knowledge}

The advantages of naivety and outsider perspectives sometimes appear in machine learning as well.  Perhaps the most famous example is AlphaGo Zero~\citep{silver2017mastering}. In contrast to standard AlphaGo, it learned to play Go only from self-play, without exposure to human games. This limitation resulted in unique and previously undiscovered strategies because it was not influenced by centuries of human play. 

More generally, more data is not always better; forcing models to operate on limited memory can improve performance in language learning~\citep{thamma2026fleeting}. ~\citet{penedo2024finewebdatasetsdecantingweb} demonstrated that a model trained on FineWeb-Edu, a well-curated 1.3T-token educational subset of the 15T-token FineWeb dataset, outperformed models trained on the full FineWeb dataset on common benchmarks like MMLU~\citep{hendrycks2021measuringmassivemultitasklanguage} and ARC~\citep{clark2018thinksolvedquestionanswering}.  The difference between these examples and the naive LLM introduced in this paper is that these past works mainly target efficiency rather than creativity.

Recently there has also been interest in training language models with a limited corpus.  Capping knowledge at a 5th grade level~\citep{li2026littlelearner} or the year 1900~\citep{hla2026machina} produces models that have interesting perspectives and unique applications, e.g.\ to test curriculum or teaching methods, or to probe whether later discoveries can be reached from earlier knowledge.  However, such models are \emph{universally} limited, whereas the models in this paper are \emph{selectively} limited.  In this paper, limiting only a portion of the model's capabilities allows the reweighting of existing information away from standard biases~\citep{zhang2026verbalized} instead of a blanket limit.  These reweightings align with approaches like novelty search~\citep{lehman2011abandoning} and quality diversity~\citep{pugh2016quality}, where the goal is to find different solutions rather than just the best one.

\subsection{Machine Unlearning} \label{sec:background_unlearning}

The main approach to achieving selectively naive LLMs in this paper is through removing data from training either so the model has never heard of the topic to begin with or to cause it to forget something it has already learned.  The former has been shown to be effective but prohibitive~\citep{o2026deep} and the latter echoes prior ``unlearning'' approaches in the field of \emph{Machine Unlearning} whose motivation is that retraining from scratch each time data needs to be scrubbed from the model would be prohibitively expensive.  The ultimate goal in these past approaches is usually to remove data for different purposes: legal, such as  privacy~\citep{jang-etal-2023-knowledge} and copyright~\citep{eldan2023whos}; safety, such as for bio-weapons~\citep{li2024wmdp} and false information~\citep{lu-koehn-2025-learn}; or alignment, like bias~\citep{dige-etal-2024-machine} and explicit content~\citep{Gandikota_2023_ICCV}.  Thus while there are many potentially relevant approaches that focus on removing training examples~\citep{liu2024rethinking}, the methods in this paper do not require such rigorous guarantees; no one will be sued if the naive model talks about an idea that turns out not to be novel.  

Other approaches steer output at inference time by adding activation vectors computed from contrastive prompts~\citep{turner2024steering, rimsky-etal-2024-steering}.  Standard unlearning methods all proved challenging in preliminary experimentation to directly apply to our domains. This limitation is directly called out by the authors of one of the most influential unlearning examples, which was able to ``unlearn'' the concept of Harry Potter~\citep{eldan2023whos}. They note that it remains uncertain whether their technique extends to non-fiction content, where there are fewer idiosyncratic terms and concepts are more embedded in broader ideas.  Real-world entities are likely to be even harder to forget; for example, forgetting Albert Einstein would touch not only on the obvious contributions to science and math, but also sociopolitical artifacts like lists of Nobel Prize winners, the Manhattan Project, and pop culture references and slang.

\section{Method} \label{sec:method}
 
This section explains both the creation of naive models and the pipeline for generating robust ideas that combines the output of the naive models with the refinement of the mentor.

\subsection{Naive Generation}
\begin{figure}
    \centering
    \includegraphics[width=1.0\linewidth]{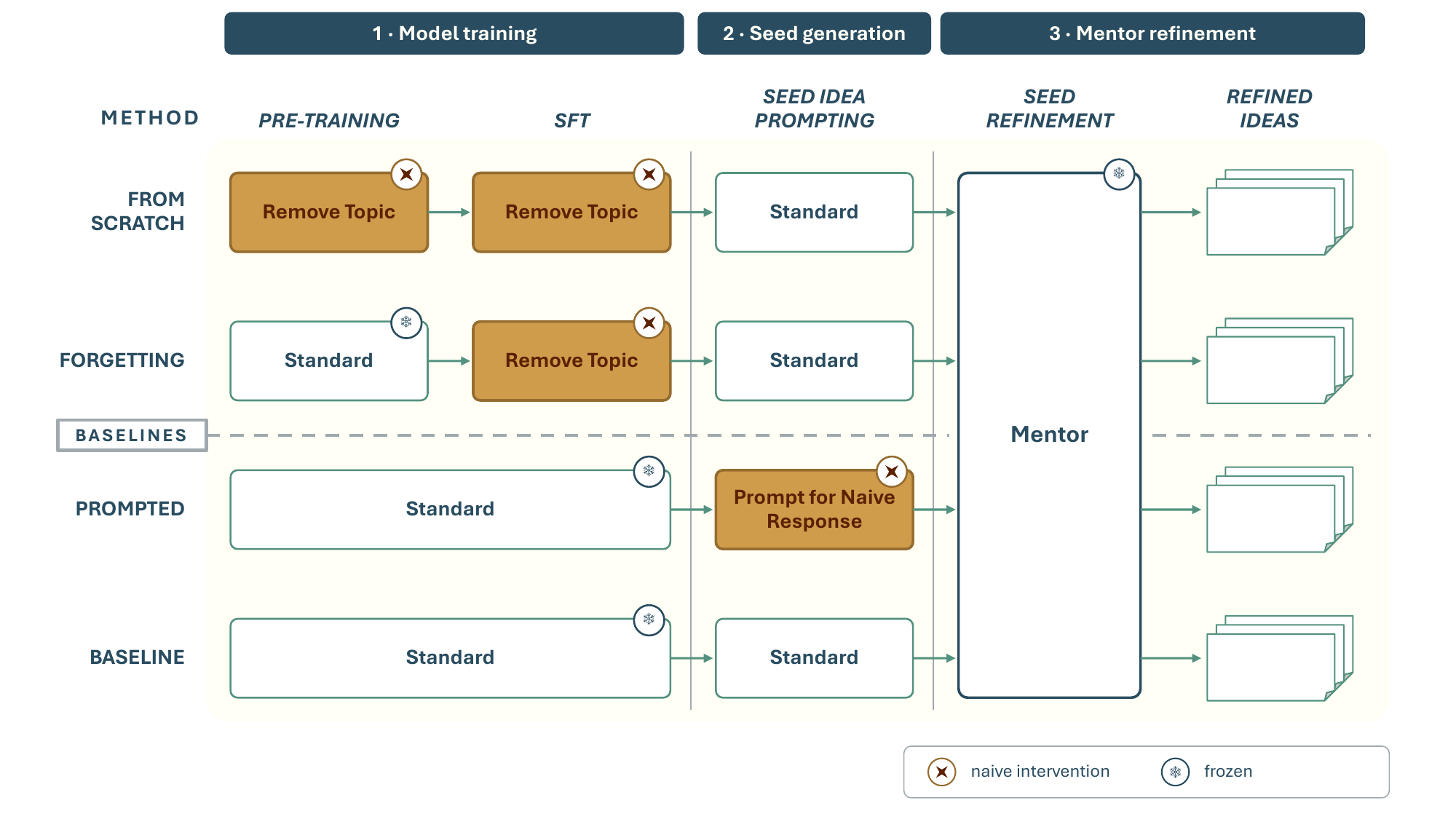}
    \caption{\textbf{Methods for obtaining naive LLMs.} There are a variety of methods to make a model unfamiliar with a topic. This paper explores \emph{training from scratch} and \emph{forgetting} (top two rows), and compares against baselines prompted for naivety and queried directly.  All approaches proceed the same way: (1) Train (or further train) a model. (2) Generate a ``seed'' idea with the model. (3) Give the seed idea to a mentor that refines the seed into a concrete proposal.  The main difference among the approaches is the point at which the topic is withheld from the model.}
    \label{fig:method}
\end{figure}

Potential approaches to generating naive output from a model vary in cost, effectiveness, and practicality.  This paper explores \emph{training from scratch} (true naive) and \emph{forgetting} (both in Figure \ref{fig:method}).

\paragraph{Training from Scratch: True Naivety.}
The purest form of naivety is to have genuinely never encountered a particular topic at all.  The only way for that to happen is to train from scratch with the topic removed from the dataset.  This approach is computationally expensive and becomes prohibitive as model size increases. It simply would not be feasible to train a new frontier-level model from scratch for every area of study. However, it is still important to explore this approach as an upper bound for what a naive perspective can achieve: what would interacting with such a model be like and how would its output be different?  For practical purposes only 1B scale models (specifically a 26-layer nanochat model from \citealp{karpathy2025nanochat}) are tested by training from scratch, but the process is general enough that it could be applied at any model size if compute were available.  The main approach is to run a filter over every piece of data before the model sees it.

Removing the topic from the dataset is achieved through a combination of basic word matching and conditional regular expressions.  For a given topic, a human and an LLM (\texttt{claude-opus-4-5-20251101}, \citealp{anthropic2025opus45}) collaboratively design the filter's terms and criteria.  The filter is blunt: if a data point is matched, the training example is discarded.  While simple in principle, such filtering is nontrivial because words and acronyms have multiple meanings, small words can fit within other words, and proper nouns like brand names can be ambiguous.  Additionally, the boundaries of a topic can be fuzzy, e.g.\ fluoride is most associated with dental care, but it also has many other uses that we would not want to remove when making a model naive to dental hygiene, especially if we want to leave open the possibility of it rediscovering the use of fluoride from scratch.  So while the filters in this paper pass through multiple rounds of refinement, they are not and probably cannot be \emph{perfect}.  They are, however, still sufficient to show significant impact with limited effect on other topics.

\paragraph{Forgetting.}

The cost of retraining a model from scratch, even at modest sizes, is prohibitive.  However, as mentioned in Section \ref{sec:background_unlearning}, there is a field dedicated to removing existing knowledge from already trained networks called \emph{machine unlearning}~\citep{liu2024rethinking}.  Such kinds of intervention usually happen after the costly pre-training step and thus may be a good middle-ground.  During preliminary research, standard unlearning techniques~\citep{liu2024rethinking,eldan2023whos} and related approaches like activation steering~\citep{turner2024steering,rimsky-etal-2024-steering} tended towards incoherence or refusal behaviors on chosen topics (which is understandable given the divergent end goals of this paper and general unlearning), so the main study in this paper does not explore them further, but more advanced or nuanced approaches may be applicable with more research.

 As a practical alternative we found an effective yet unconventional approach to intentional forgetting based on a phenomenon LLM training typically tries to avoid: \emph{catastrophic forgetting}~\citep{mccloskey1989catastrophic}.  The key phenomenon is when a model loses knowledge or a skill during extended training (typically post-training) once it is no longer exposed to the original knowledge~\citep{luo2025empirical}.  Because the goal of training naive LLMs is different from typical unlearning (e.g.\ there are no legal or ethical repercussions to retaining the data), this approach offers the ability to gradually erode information and connections in a way that loses fine details but may retain surface-level information. This kind of partial forgetting is reminiscent of a real naive person who may have casually heard of a topic, but does not understand it in depth.

 Practically, forgetting is implemented with a model that has gone through a standard pre-training process by applying the filter described above to \emph{only} the SFT dataset.  To achieve the forgetting, more SFT epochs than normal are applied and thus training continues until performance on a topic-specific dataset falls below a threshold, typically two to three epochs beyond the standard training recommendations for a 7B model.  However, anecdotally, even after ten or more additional epochs of training the model remained coherent, unlike the other unlearning approaches attempted.

\paragraph{Prompting.}
As a control, we can also simply ask the models to be naive.  Asking a model to adopt a persona or take a specific viewpoint is a common prompt-engineering technique~\citep{lutz-etal-2025-prompt}. While this approach offers a low-cost solution, a challenge is that LLMs cannot reliably suppress prior knowledge on instruction~\citep{li2026simulated} and only imperfectly emulate weaker models~\citep{vanderweij2024sandbagging}; simulating true naivety is difficult when you are already an expert.  To implement this approach, a baseline version of the model that was trained without any filtering has its prompts prepended with, ``Pretend you have never heard of \verb|{$topic_description}| and answer the following question as such:''.  The other control in this paper is to simply ask the baseline model the question directly, creating a baseline response to compare naive approaches against.

\subsection{Mentors}
While the naive approaches can generate interesting \emph{seed} ideas, almost by definition, their lack of knowledge about the topic means their ideas may need refinement, which  raises the idea of a \emph{mentor} model.  By giving a strong mentor model the naive model’s \emph{seed} idea, the mentor can be inspired 
to fill in holes or inconsistencies in the naive model’s initial thoughts.  In this paper, Gemini 3.1 Pro~\citep{googledeepmind2026gemini31pro} refines the seeds, although any sufficiently capable model could work.

The process for mentoring is straightforward (Figure \ref{fig:method}): a model generates a seed response to the initial question and then that seed is passed on to the mentor model with framing text (Appendix \ref{app:mentor_prompt}).  The mentor is specifically instructed to follow the seed and it is later judged on this alignment; otherwise, strong models were found to discard an unconventional seed and invent their own, ``better'' approach whole cloth.  The mentor is also instructed to develop only the most interesting idea if a seed response contains multiple possible directions. Otherwise the mentors would again gravitate towards more standard, in-distribution options if they have to pick among multiple options. Unchecked, the mentors always preferred to steer the discussion to more conventional approaches, but when they are forced to engage with the naive seeds they can refine them into unique final ideas.

\section{Experiments} 

In this paper, naive models are generated across a variety of topics (one model per topic, per size) and asked to solve problems relevant to that same topic.  This section details the models trained and tested, how topics are generated, and how the responses are evaluated.

\subsection{Models}

Naive models are explored at 1B and 7B sizes.  Models follow their recommended training recipes (see Appendix \ref{app:training_configs} for details) with the only change being the data filtering described in Section \ref{sec:method}.

The smallest model is a depth 26 nanochat~\citep{karpathy2025nanochat} (roughly 1B parameters) that is run with pre, mid, and SFT training; the optional RL step is not applied. The nanochat model was chosen for its open framework and fast iteration speed.  For 7B experiments, the process starts with \texttt{allenai/Olmo-3-1025-7B}, which is already pretrained on an unfiltered dataset and is then further trained with its standard Instruct SFT dataset (Dolci-Instruct-SFT)~\citep{olmo2025olmo3} and no further training stages.  The Olmo family was (again) chosen for the open framework and training process.  \emph{Gemini 3.1 Pro} serves as the mentor.  The fact that Gemini is a different model family from the others benefits it in the role of a mentor because in some cases LLMs are biased to their own output~\citep{panickssery2024llm}. That said, any sufficiently capable frontier model could take on the role.

\subsection{Naive Topics}
Because of the manual curation of training data filters and questions, this study is limited to four topics. It may be possible to automate the filtration process in the future, but initial tests suggest that topic fuzziness and scope benefit from human oversight to be effective.  The four topics are \emph{dental hygiene}, \emph{neuroevolution}, \emph{vaccination}, and \emph{backpropagation}.  They were chosen to cover a breadth of complexity levels and levels of required technical knowledge. For example, dental hygiene is common knowledge that most people would be expected to have some familiarity with, whereas neuroevolution~\citep{stanley2019designing} is a somewhat niche machine learning field (that the authors are familiar with and qualified to make nuanced judgments on).

However, importantly, all of these topics have open problems and challenges where fresh eyes can potentially help.  For each topic, in addition to the filter terms, a set of topic-specific problem-solving questions is generated (Appendix \ref{app:topic_prompts}).  These are roughly of the form, ``I am experiencing \verb|{problem in the field}|.  Can you help me solve it?'' or ``I need help with \verb|{thing the field helps solve}|.  Can you help me invent something to solve it?'' The specific formulation varies by the topic, but the goal is to elicit a helpful seed response from the model that can be practically refined by the mentor. For that reason, simple definitional questions like, ``What is a toothbrush?'' are avoided; the responses to such questions can be novel or even entertaining, but there is nothing for the mentor to refine.

\subsection{Evaluation}

Seed ideas and mentor refinements are evaluated in several ways: an LLM judge produces qualitative metrics, embedding distance further assesses distributional shifts quantitatively, and distributions are also analyzed qualitatively through lexical features.  Multiple metrics are helpful because no single measure is reliable on its own: LLM judges can be biased, embedding distances can conflate content with style, and lexical features show what differs but not whether it matters.

Idea novelty is a key inspiration for exploring naive output; in real life, approaching a problem with ``fresh eyes'' is a great way to break out of typical patterns.  Yet a subjective metric like ``novelty'' is notoriously difficult to measure precisely~\citep{lamb2018evaluating}.  LLM judges serve as an emerging option~\citep{zheng2023judging} but still have issues that must be mitigated. In this paper, judges rank all responses (seeds and refinements) on three criteria on a scale of 1--10 (Appendix \ref{app:judge_prompts} includes rubrics): novelty, utility, and feasibility.  The latter two are meant as a sanity check on the first; it is easy to come up with a novel but useless or impractical idea.  It may even be expected that the naive models do so because they lack technical grounding. The job of the mentor is to ensure the proposal is sound across all the metrics, although the model is not explicitly told about the metrics.  Mentor refinements are additionally assessed on faithfulness to the seed prompt to avoid the mentor substituting a different idea if it does not like the seed.  To balance cost and effectiveness (cheaper but less capable judges struggled with the nuanced judgments required), \texttt{GPT-5.6 Luna}~\citep{openai2026gpt56luna} is always the judge. An additional benefit is that Luna belongs to a different model family than all the other models in the experiments and does not suffer from in-family bias mentioned earlier.

Embedding distances are a quantitative measure of how much a naive model's response deviates from a non-naive baseline response of the same model type.  All responses are encoded with \texttt{Qwen3-Embedding-0.6B}~\citep{zhang2025qwen3embedding}.  Because the prompts are all related to the same topic, embeddings are residualized before comparing to remove the encoded topic description by subtracting the topic direction (cf.\ iterative nullspace projection~\citep{ravfogel2020null}) and then unit-normalized.  The seed-paired average cosine distance, \begin{equation}
    \bar{d}_{\cos} = \frac{1}{S}\sum_{s=1}^{S}\left(1 - \frac{\mathbf{u}_s \cdot \mathbf{v}_s}{\lVert \mathbf{u}_s \rVert \, \lVert \mathbf{v}_s \rVert}\right),
\end{equation} between treatment responses (trained naive, prompted naive, and baseline) is computed, where $S$ is the number of random seeds, and $\mathbf{u}_s$ and $\mathbf{v}_s$ are the embeddings produced under seed $s$ by the two conditions being compared.  They are compared with PERMANOVA with $9$,$999$ label permutations to compute group differences~\citep{andersion2001permanova}.  To determine which method is further from the baseline, a one-sided relative MMD test~\citep{bounliphone2016testrelativesimilaritymodel} is applied.  Because UMAP and t-SNE 2D projections can distort distances and cluster structure~\citep{wattenberg2016use, chari2023specious}, the embeddings are instead visualized by projecting onto a single linear direction via group means, following the difference-of-means projections used to assess class separability in LLM activations~\citep{braun2025understanding}.  Comparing the seed ideas to their mentor refinements would also be interesting in principle, but the refinement is generated by a different model family and size, which typically dominate embedding differences and so these comparisons are not performed.

Because embeddings can conflate things like content and style~\citep{wegmann-etal-2022-author} and single-vector embeddings have fundamental limits in representing fine-grained distinctions that standard benchmarks do not reveal~\citep{weller2026theoretical}, lexical features are also explored.  Specifically through the ``Fightin' words'' approach~\citep{monroe2008fightin} that looks at the distinctive terms from each group and ranks them by the z-scored weighted log-odds ratio with an informative Dirichlet prior.  Positive values are over-represented in the named group and negative values are under-represented with shared vocabulary shrunk toward zero.  This analysis is applied to the mentor refinements to compare the final ideas.

\section{Results}

Results are shared first for truly naive 1B models that are trained from scratch, followed by an examination of forgetting in larger 7B models.  For each model size and topic explored, a new naive LLM is trained and 32 seed ideas are generated for each of three topic-dependent prompts from the naive, baseline, and prompted naive (i.e.\ the baseline model with instructions to pretend it does not know the topic) models.  Then these seeds are refined by the mentor (frontier model) and both the seed and refinement are evaluated.

\subsection{Truly Naive 1B Models}

\begin{figure}[H]
    \centering
    \begin{subfigure}[ht]{0.48\textwidth}
        \centering
        \includegraphics[width=1.0\linewidth]{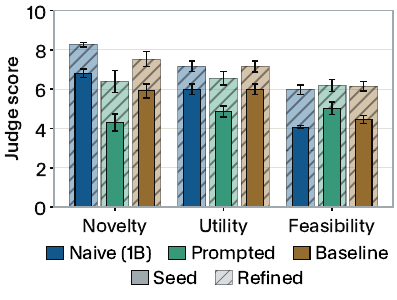}
        \caption{Dental Hygiene Judge Scores}
        \label{fig:1b_dental_judge}
    \end{subfigure}
    \begin{subfigure}[ht]{0.48\textwidth}
        \centering
        \includegraphics[width=1.0\linewidth]{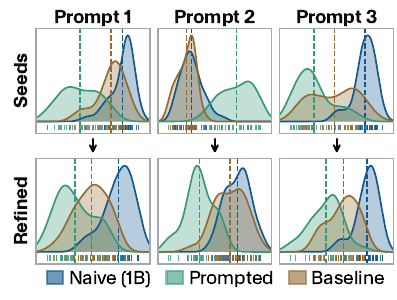}
        \caption{Dental Hygiene Response Embeddings}
        \label{fig:1b_dental_embeddings}
    \end{subfigure}
    \caption{\textbf{1B naive models outperform baselines on dental hygiene ideas.}  The judge scores are shown in (a), averaged across all prompts for a topic. All seed scores are shown in solid colors and mentor refinement scores are hashed.  Bars show mean score across all prompts; error bars show 95\% CI.  For the relatively simple topic of dental hygiene, true naive models produce more novel, but less feasible seed ideas (a).  After mentor refinement, feasibility equalizes and novelty further improves. The embedding images (b) reflect the PERMANOVA and MMD analyses. They are represented as 1-D projections of individual outputs (bottom ticks) through the three group means (dashed lines) following from difference-of-means projections in analyzing LLM class separation~\citep{braun2025understanding}. Curves are per-group densities with less overlap indicating greater separation.  They demonstrate significant distances between the naive (blue) and both baseline approaches for both initial seeds and refinements, although in some cases (e.g.\ as in prompt 2) the baseline and naive overlap substantially.}
    \label{fig:1b_dental}
\end{figure}

At the 1B scale, it is possible to test truly naive models by excluding specific topics from the training set.  Results are compared to a baseline model that was trained without any topic filtering and to the same baseline simply prompted to be naive.  In the dental hygiene domain, the from-scratch naive model seeds are scored as significantly more novel (Figure \ref{fig:1b_dental_judge}) than the baseline and prompted models ($p<0.001$ for both), although they are also rated significantly less feasible than both ($p<0.001$).  When the mentor is applied, the difference in feasibility is removed (i.e.\ the naive proposals are made as scientifically plausible as baseline, standard proposals) and the from-scratch naive approaches become even more novel, retaining their significant lead ($p<0.001$).  Accordingly, the from-scratch naive model produces significantly different outputs than both the baseline and prompted naive models ($p<0.001$ for both seeds and refinements, Figure \ref{fig:1b_dental_embeddings}) overall, although the distributions specifically for prompt 2 overlap heavily. The implication is that training from scratch can indeed affect model output on a specific topic.  Also, the mentor is able to enhance the novelty while finding practical use cases that the naive model is unaware of. 

\begin{figure}
    \centering
    \begin{subfigure}[ht]{0.48\textwidth}
        \centering
        \includegraphics[width=1.0\linewidth]{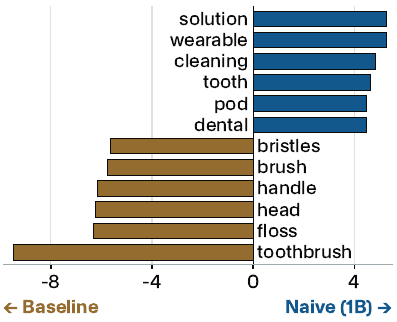}
        \caption{Baseline vs.\ Naive Differential Word Usage}
        \label{fig:1b_dental_bvn}
    \end{subfigure}
    \begin{subfigure}[ht]{0.48\textwidth}
        \centering
        \includegraphics[width=1.0\linewidth]{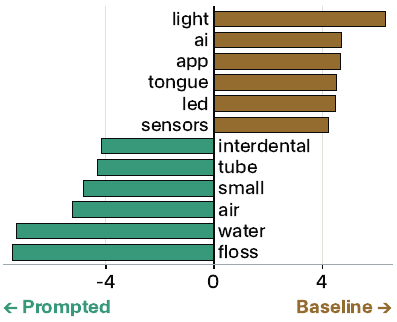}
        \caption{Baseline vs.\ Prompted Differential Word Usage}
        \label{fig:1b_dental_bvp}
    \end{subfigure}
    \caption{\textbf{Differential Word Usage in 1B Model Refinements.}  This figure shows the distinctive terms used in each model's refinements for the given pair ranked by the $z$-scored weighted log-odds ratio with significance at $|z| > 1.96$. Shared topic vocabulary is shrunk toward zero, so the listed terms reflect what differs between groups.  The naive refinements focus on the methods (cleaning, wearable), while the baseline model sticks to standard dental cleaning implements.  Comparing the prompted naive model to the baseline, most common cleaning implements (e.g.\ toothbrush) do not appear (except floss, which did not appear in (a) either) and are thus heavily shared between both the prompted and baseline models. All such comparisons are in Appendix \ref{app:word_usage}.}
    \label{fig:1b_dental_tornados}
\end{figure}

However, the mentor's refinement is not without a cost; while the differences in embedding space remain significant in this model, the mentor can reduce them.  One reason for this reduction is that the mentor model \emph{actually knows} about the topic at hand and is thus free to incorporate common terms and ideas, e.g.\ ``standard dental practice relies on nylon bristles and waxy floss...''  Additionally, the scientific grounding can remove some of the more fantastical or speculative elements, e.g.\ vague appeals to ``AI-powered'' sensing and ``advanced nanotechnology'' in 1B naive seeds are typically dropped while the underlying physical concept (e.g.\ an in-mouth device or micro-saws) is retained and given a concrete mechanism.  Furthermore, the refined responses are all produced by the same, untreated frontier model, which likely reduces any stylistic or syntactic deltas.  Ultimately, the models still produce different ideas, as shown by the judge scores for the mentor refinements (Figure \ref{fig:1b_dental_judge}) and by examining the lexical features of the output (Figure \ref{fig:1b_dental_tornados}).  For example, the prompted and baseline both heavily reference common dental implements (e.g.\ floss and toothbrushes) while the naive models focus more on techniques and methods.  %

Qualitatively, the baseline model tends to propose extensions to existing dental hygiene products like a floss holder or to add features like Bluetooth to a toothbrush. The prompted naive model either gives definitional answers (e.g.\ ``a toothbrush is...'') or sometimes even ignores the ``pretend you don't know'' instructions entirely.  In contrast, the from-scratch naive model proposes more radical inventions like a mouth-cleaning drone, or a ``tongue stone'' that cleans teeth hands-free by being pushed around with your tongue.  When given to the mentor the novel ideas are further elaborated and made more practical: in the case of the ``tongue stone'' the mentor suggests that the ``coating is microporous and infused with nano-hydroxyapatite (nHAp),'' grounding vague concepts in the seed to more concrete concepts that the naive model could not even define if asked about them directly. The baseline and prompted naive refinements are also lifted by the mentor, but they do not provide much to build on. As a consequence, there is only so much the mentor can do with generic suggestions.  Appendix \ref{app:examples} has more examples.

\begin{figure}
    \centering
    \begin{subfigure}[ht]{0.48\textwidth}
        \centering
        \includegraphics[width=1.0\linewidth]{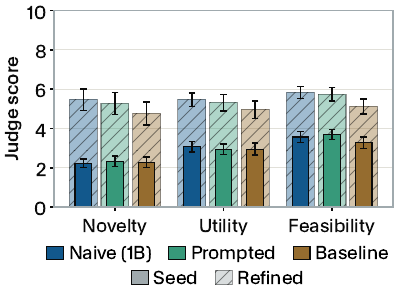}
        \caption{Neuroevolution Judge Scores}
        \label{fig:1b_neuro_judge}
    \end{subfigure}
    \begin{subfigure}[ht]{0.48\textwidth}
        \centering
        \includegraphics[width=1.0\linewidth]{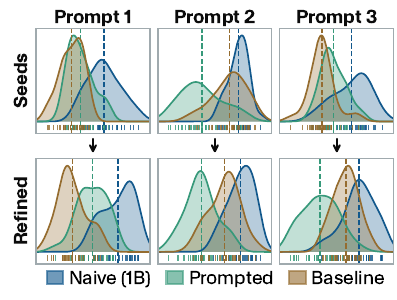}
        \caption{Neuroevolution Response Embeddings}
        \label{fig:1b_neuro_embeddings}
    \end{subfigure}
    \caption{\textbf{1B naive models do not work for complex topics.}  Results are presented in the same format as in Figure \ref{fig:1b_dental}. For more complex topics like neuroevolution here, all the methods end up producing similarly rated (a) generic ideas that map similarly in embedding space after refinement (b).  A stronger naive model allows the topic to be explored more effectively (later shown in Figure \ref{fig:7b_neuro}). Bars show mean score across all prompts; error bars show 95\% CI.}
    \label{fig:1b_neuro}
\end{figure}

Results are different for the topic of neuroevolution. Here, the approaches are largely indistinguishable, i.e.\ there is no significant difference in judge scores (Figure \ref{fig:1b_neuro_judge}) or embedding space (Figure \ref{fig:1b_neuro_embeddings}).  The problem is first that the base nanochat model does not know much about the niche topic of neuroevolution to begin with, so the difference among them is relatively small. Second, the model's 1B capacity limits its ability to engage with complex topics in general, damping the possible advantage of being naive in such an area.  All 1B variants in this domain tend to give generic machine learning advice that the mentor does its best to apply to neuroevolution, occasionally resulting in useful insights, but typically not pushing innovation.

Overall, the 1B true naive model tests hint at the potential for naive networks to provide seeds for alternative ideas different from what base networks would normally offer, but the limited intelligence at 1B constrains the breadth of domains where this potential can be applied.

\subsection{Forgetting for Naive 7B Models}

The limited capabilities of the 1B model in the neuroevolution domain imply that larger models may be necessary to ideate about more complex topics.  However, as models get larger, pretraining time increases dramatically. Therefore, for 7B models, the forgetting method serves as the practical alternative to training from scratch.  The increased capabilities of the larger model also motivate trying a wider range of topics: dental hygiene, neuroevolution, vaccination, and backpropagation.

\begin{figure}
    \centering
    \begin{subfigure}[ht]{0.48\textwidth}
        \centering
        \includegraphics[width=1.0\linewidth]{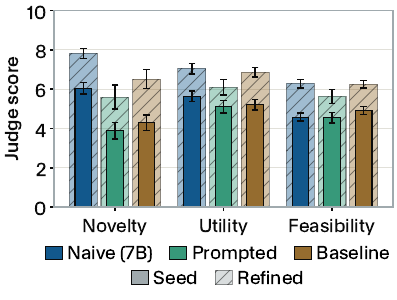}
        \caption{Dental Hygiene Judge Scores}
        \label{fig:7b_dental}
    \end{subfigure}
    \begin{subfigure}[ht]{0.48\textwidth}
        \centering
        \includegraphics[width=1.0\linewidth]{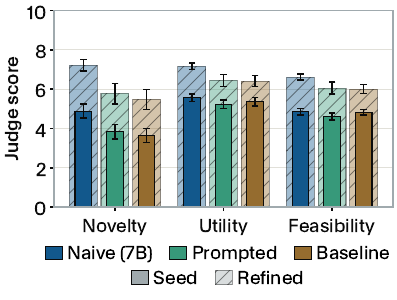}
        \caption{Neuroevolution Judge Scores}
        \label{fig:7b_neuro}
    \end{subfigure}
    \begin{subfigure}[ht]{0.48\textwidth}
        \centering
        \includegraphics[width=1.0\linewidth]{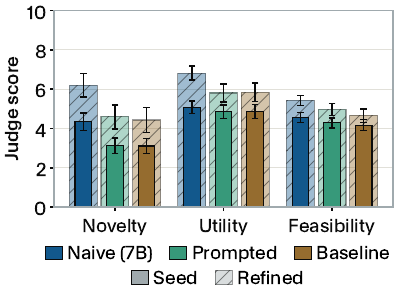}
        \caption{Vaccination Judge Scores}
        \label{fig:7b_vaccines}
    \end{subfigure}
    \begin{subfigure}[ht]{0.48\textwidth}
        \centering
        \includegraphics[width=1.0\linewidth]{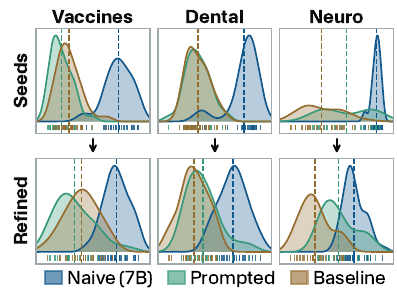}
        \caption{Per-Topic Embeddings}
        \label{fig:7b_embedding}
    \end{subfigure}
    \caption{\textbf{7B models made to forget a topic outperform baselines on a variety of topics.}  Results are reported in the same format as in Figure \ref{fig:1b_dental}.  At this larger model size, naive forgetting seeds and refinements  generate more novel and useful solutions with competitive feasibility scores (a-c).  The naive seeds also tend to map to different parts of the embedding space compared to the baseline models, but the overlap increases for refinements as the mentor grounds them (d).  See Appendix \ref{app:add_embed} for all embeddings.  Bars show mean scores across all prompts; error bars show 95\% CI.}
    \label{fig:7b_models}
\end{figure}

Across all domains except backpropagation, the forgetting naive models produce more novel seed ideas that are refined to even more novel final proposals ($p<0.001$ for seeds and refinements versus other methods, Figure \ref{fig:7b_models}a-c).  Utility remains similar among seeds (except in a few cases, e.g.\ 7B Dental naive seeds are significantly higher, $p<0.01$), but when refined by the mentor the naive ideas also have significantly higher utility (at least $p<0.01$) except for the dental baseline.  

\begin{figure}
    \centering
    \begin{subfigure}[ht]{0.48\textwidth}
        \centering
        \includegraphics[width=1.0\linewidth]{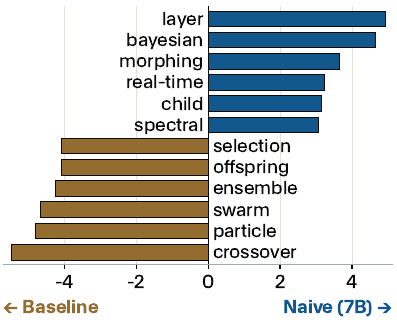}
        \caption{Baseline vs.\ Naive Differential Word Usage for Neuroevolution}
        \label{fig:7b_neuro_nvb}
    \end{subfigure}
    \begin{subfigure}[ht]{0.48\textwidth}
        \centering
        \includegraphics[width=1.0\linewidth]{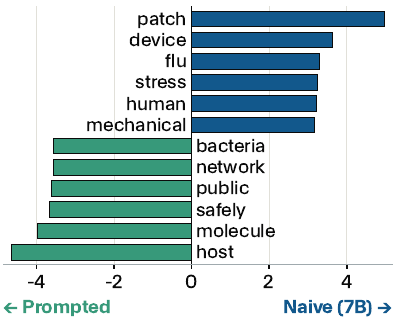}
        \caption{Baseline vs.\ Prompted Differential Word Usage for Vaccines}
        \label{fig:7b_vaccines_nvp}
    \end{subfigure}
    \caption{\textbf{Differential Word Usage in 7B Model Refinements.}  Results are reported in the same format as Figure \ref{fig:1b_dental_tornados}.  The 7B naive forgetting models follow the trend in the 1B models of focusing on fewer in-topic words and instead describe more about methods or adjacent approaches when compared to the baseline and prompted methods.  For all such comparisons see Appendix \ref{app:word_usage}.}
    \label{fig:7b_tornados}
\end{figure}

Embedding distances generally show significant separation between baseline and naive models for seeds ($p<0.05$, see Figure \ref{fig:7b_embedding} for examples) but only some of the refinements have significant embedding differences (e.g.\ $p<0.01$ for dental) over all prompts.  This standardizing of representations through refinement was noted in the 1B model, but here the effect is more prominent.  Despite that, qualitative analysis of word likelihood (Figure \ref{fig:7b_tornados}) and the judge scores strongly imply that the naive model is exploring unique ideas, such as in the 7B neuroevolution naive model (Figure \ref{fig:7b_neuro_nvb}). That model focuses on adjacent fields like ``Bayesian'' or seemingly off-topic ideas like ``spectral,'' though it is possible that the embedding models have difficulty separating nuance within specific topics.  Additionally, the naive seeds and responses are consistently further from the baseline model than the prompted seeds are.

\begin{figure}
    \centering
    \begin{subfigure}[ht]{0.48\textwidth}
        \centering
        \includegraphics[width=1.0\linewidth]{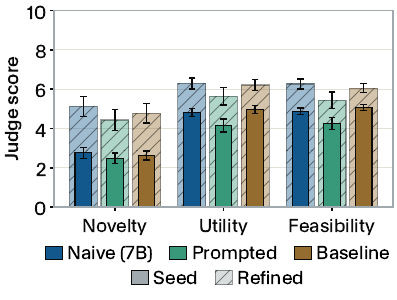}
        \caption{Backpropagation Judge Scores}
        \label{fig:7b_backprop_judge}
    \end{subfigure}
    \begin{subfigure}[ht]{0.48\textwidth}
        \centering
        \includegraphics[width=1.0\linewidth]{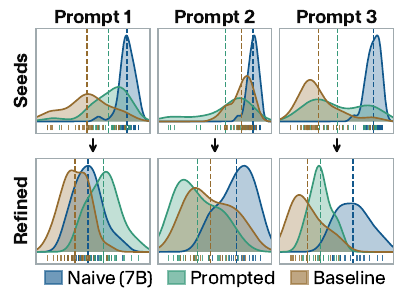}
        \caption{Backpropagation Response Embeddings}
        \label{fig:7b_backprop_embed}
    \end{subfigure}
    \caption{\textbf{7B naive models fail to gain an advantage in the backpropagation domain.} The same format appears here as in Figure \ref{fig:1b_dental}. Much like the 1B models failing to provide good seeds to neuroevolution, the 7B models struggle with backpropagation, with no significant difference in judge scoring (a) and all responses map to a similar embedding space after refinement (b).  Bars show mean score across all prompts; error bars show 95\% CI.}
    \label{fig:7b_backprop}
\end{figure}

In contrast to these positive results for the naive (from forgetting) 7B networks, the results in the backpropagation domain at 7B look much like the neuroevolution results at 1B, with no significant difference in scores for seeds or refinements (Figure \ref{fig:7b_backprop_judge}).  The embeddings (Figure \ref{fig:7b_backprop_embed}) do, however, show significant differences ($p<0.001$) in the seeds, so the naive model does produce different solutions, just not particularly good ones.  Overall the implication is that an even larger naive model may be required to be able to grapple effectively with complicated machine learning topics.

Qualitatively, responses at 7B differ from the 1B model responses in several ways.  First, due to the model size, the results tend to be more expressive and more grounded in reality, especially for scientific topics.  The forgetting-naive models also have not completely forgotten the topics; they still make reference to common items in the domains but will make nuanced changes that trigger the mentor to think about the problem in a new light.  For example, a vaccine seed idea is a patch that monitors viral exposure levels and the mentor refined it into a ``CRISPR-Cas12/13 biosensing lapel pin that logs exposure events to measure real-world vaccine efficacy and guide targeted immunizations.''  The combination of the fuzzy nature of topic boundaries and the strength of the pre-training examples may make it difficult to completely ``forget'' a topic in this manner, but the forgetting approach achieves the desired result of novel outputs by approaching naivety through the degradation of prior knowledge.  Appendix \ref{app:examples} shares more examples.

Overall, the trend is consistent: whether through training from scratch or through forgetting, naive models consistently exhibit significantly more novel proposals than both baseline models and baseline models prompted to be naive.  These results are generally preserved through mentor refinement, which also brings up feasibility and utility scores, though the novelty advantage may sometimes narrow (while remaining significant).  At the same time, small naive models lose their advantage when the topic is too complex for their general level of competence.  Nevertheless, the overall trend suggests the breadth of benefit of naive models should only grow as larger models are made naive.

\section{Discussion}

The argument is rarely made in machine learning that knowing less, having less organized understanding, or being deprived of relevant data can be an advantage, especially when the deficit is precisely in the area being evaluated.  Indeed, in most problems and benchmarks, being naive or ignorant is likely a significant disadvantage.  However, results with 1B and especially 7B models in this paper hint that under the right conditions problems involving \emph{creativity} may be an exception. It might sometimes be useful for creative purposes to know less about the area you are going into.

Furthermore, it appears difficult to \emph{fake} being naive; simply asking the 1B or 7B baseline model to take a naive perspective fails to compete with a model that went through a genuine forgetting process.  This result makes intuitive sense; if having fresh eyes really does confer a creative benefit, then if it could work, many a one-hit wonder would simply pretend to be starting over again, but it clearly is not that easy.

The hint that knowing less can be advantageous cuts against longstanding assumptions in the field and undercuts the ``knowing about everything'' large model archetype that currently dominates both frontier inspiration and discourse. Yet while this less-is-more insight may initially seem surprising, it aligns with reasonable and longstanding intuitions about the value of fresh eyes~\citep{azoulay2019funeral}. Unchained from dogma and freed from entrenched assumptions, it makes sense that a newcomer can pave a new path forward.

However, that newcomer must still be \emph{generally} intelligent.  One potential point of confusion would be to conflate \emph{being naive in a particular area} with \emph{being generally less capable}. Clearly overall diminished faculties are not the same advantage as coming to a field with fresh eyes and an otherwise vibrant mind, like a promising young student or novel entrant to a field.

This critical yet subtle distinction between a naive but gifted mind versus a generally impaired mind highlights why it was critical to perform baseline comparisons with models of the same overall capacity.  That is, the naive (from forgetting) 7B model was compared to other 7B baselines (a regular 7B model and one asked to ``pretend'' to be naive).  Even with a mentor, it is important to control for the capacity of the naive LLM because a mentor network that is genuinely helping another network refine \emph{its own} ideas is intrinsically constrained by the capacity or intelligence of the network it is assisting.

The nature of conventional LLM training makes it difficult to straightforwardly probe the implications of fresh eyes in a particular area.  
Training an entire LLM from scratch with one field of knowledge cut from its dataset is prohibitive (which is why we only attempted it at the 1B level) and we struggled to apply machine unlearning techniques to mature networks for this purpose.  The forgetting method we introduce in this paper, which intentionally capitalizes on catastrophic forgetting, may not be the last word on obtaining naive LLMs, but it works well enough to suggest their intriguing potential.

In fact, the most important contribution of this work is likely not the method of forgetting, but rather the provocation of results suggesting that knowing less about something can be an advantage.  That insight is likely a stepping stone to many novel ideas and experiments later to be conceived.

The mentor networks add a corollary to the main insight: the relative advantage of fresh eyes can be amplified with assistance.  Note that the mentor in this paper is not a clean metaphor for mentors in real life, because it is \emph{both} significantly more capable in general (being a frontier model versus the 7B model mentee) and also an expert in the area of ignorance of the naive LLM.  Human apprentice-mentor relationships are likely to differ in the sense that the mentor is not necessarily vastly more intelligent in general.  However, the results with mentors still help to highlight the fact that what might start out looking like only a subtle advantage from fresh eyes can end up greatly amplified under the right guidance.

A natural and intriguing question for the future is what a naive frontier-level LLM might be like.  What would 1T or larger networks be capable of imagining given a blank slate in a particular area?  Achieving such networks takes significantly more compute than 7B experiments in this paper, and also might require careful elaboration and calibration of the forgetting technique developed here that worked on much smaller networks.  In the end, the best directions to take the central insight are unpredictable, but our hope is that the hint of potential here will inspire investigations in the future that heretofore would have been dismissed as unprincipled; one can no longer claim with full confidence that more data and more knowledge is always better, even when the deficit is in the domain of interest itself.

\section{Conclusion}

 Purposefully restricting an LLM's knowledge of a topic can mimic a human coming to a field with ``fresh eyes,'' unencumbered by existing bias or ways of approaching a problem. %
 The naive models in this paper regularly outperformed baseline and naive prompted models in novelty and utility metrics, and embedding analysis showed that naive approaches produce a unique type of novelty.  Pairing these naive models with a strong mentor generally boosted all aspects of proposals, including elevating feasibility to match baseline approaches. %
 Contrary to scaling laws, this paper demonstrates a case where less knowledge can actually be more powerful, which merits future explorations across greater model sizes.  

\bibliography{references}

\begin{thebibliography}{68}
\providecommand{\natexlab}[1]{#1}
\providecommand{\url}[1]{\texttt{#1}}
\expandafter\ifx\csname urlstyle\endcsname\relax
  \providecommand{\doi}[1]{doi: #1}\else
  \providecommand{\doi}{doi: \begingroup \urlstyle{rm}\Url}\fi

\bibitem[Alon et~al.(2026)Alon, Bloom, Gowers, Litt, Sawin, Shankar, Tsimerman, Wang, and Wood]{alon2026remarks}
Noga Alon, Thomas~F. Bloom, W.~T. Gowers, Daniel Litt, Will Sawin, Arul Shankar, Jacob Tsimerman, Victor Wang, and Melanie~Matchett Wood.
\newblock Remarks on the disproof of the unit distance conjecture, 2026.

\bibitem[Anderson(2001)]{andersion2001permanova}
Marti~J. Anderson.
\newblock A new method for non-parametric multivariate analysis of variance.
\newblock \emph{Austral Ecology}, 26\penalty0 (1):\penalty0 32--46, 2001.
\newblock \doi{10.1111/j.1442-9993.2001.01070.pp.x}.
\newblock URL \url{https://onlinelibrary.wiley.com/doi/abs/10.1111/j.1442-9993.2001.01070.pp.x}.

\bibitem[{Anthropic}(2025)]{anthropic2025opus45}
{Anthropic}.
\newblock Claude opus 4.5 system card.
\newblock System card, November 2025.
\newblock URL \url{https://www.anthropic.com/claude-opus-4-5-system-card}.
\newblock API model snapshot \texttt{claude-opus-4-5-20251101}. Accessed 2026-09-10.

\bibitem[Azoulay et~al.(2019)Azoulay, Fons-Rosen, and Graff~Zivin]{azoulay2019funeral}
Pierre Azoulay, Christian Fons-Rosen, and Joshua~S. Graff~Zivin.
\newblock Does science advance one funeral at a time?
\newblock \emph{American Economic Review}, 109\penalty0 (8):\penalty0 2889--2920, August 2019.
\newblock ISSN 0002-8282.
\newblock \doi{10.1257/aer.20161574}.

\bibitem[Bounliphone et~al.(2016)Bounliphone, Belilovsky, Blaschko, Antonoglou, and Gretton]{bounliphone2016testrelativesimilaritymodel}
Wacha Bounliphone, Eugene Belilovsky, Matthew~B. Blaschko, Ioannis Antonoglou, and Arthur Gretton.
\newblock A test of relative similarity for model selection in generative models.
\newblock In \emph{International Conference on Learning Representations}, 2016.
\newblock URL \url{https://arxiv.org/abs/1511.04581}.

\bibitem[Braun et~al.(2025)Braun, Eickhoff, Krueger, Bahrainian, and Krasheninnikov]{braun2025understanding}
Joschka Braun, Carsten Eickhoff, David Krueger, Seyed~Ali Bahrainian, and Dmitrii Krasheninnikov.
\newblock Understanding (un)reliability of steering vectors in language models, 2025.
\newblock URL \url{https://arxiv.org/abs/2505.22637}.
\newblock ICLR 2025 workshop paper.

\bibitem[Chari \& Pachter(2023)Chari and Pachter]{chari2023specious}
Tara Chari and Lior Pachter.
\newblock The specious art of single-cell genomics.
\newblock \emph{PLOS Computational Biology}, 19\penalty0 (8):\penalty0 e1011288, 2023.

\bibitem[Clark et~al.(2018)Clark, Cowhey, Etzioni, Khot, Sabharwal, Schoenick, and Tafjord]{clark2018thinksolvedquestionanswering}
Peter Clark, Isaac Cowhey, Oren Etzioni, Tushar Khot, Ashish Sabharwal, Carissa Schoenick, and Oyvind Tafjord.
\newblock Think you have solved question answering? try arc, the ai2 reasoning challenge, 2018.
\newblock URL \url{https://arxiv.org/abs/1803.05457}.

\bibitem[Cui et~al.(2026)Cui, Lin, Wu, and Evans]{cui2026aging}
Haochuan Cui, Yiling Lin, Lingfei Wu, and James~A. Evans.
\newblock Aging and the narrowing of scientific innovation.
\newblock \emph{Science}, 392\penalty0 (6798):\penalty0 588--591, 2026.
\newblock \doi{10.1126/science.ady8732}.

\bibitem[Dige et~al.(2024)Dige, Arneja, Yau, Zhang, Bolandraftar, Zhu, and Khattak]{dige-etal-2024-machine}
Omkar Dige, Diljot Arneja, Tsz~Fung Yau, Qixuan Zhang, Mohammad Bolandraftar, Xiaodan Zhu, and Faiza~Khan Khattak.
\newblock Can machine unlearning reduce social bias in language models?
\newblock In \emph{Proceedings of the 2024 Conference on Empirical Methods in Natural Language Processing: Industry Track}, pp.\  954--969, Miami, Florida, US, November 2024. Association for Computational Linguistics.
\newblock \doi{10.18653/v1/2024.emnlp-industry.71}.
\newblock URL \url{https://aclanthology.org/2024.emnlp-industry.71/}.

\bibitem[Eldan \& Russinovich(2023)Eldan and Russinovich]{eldan2023whos}
Ronen Eldan and Mark Russinovich.
\newblock Who's harry potter? approximate unlearning in llms, 2023.
\newblock URL \url{https://arxiv.org/abs/2310.02238}.

\bibitem[Gandikota et~al.(2023)Gandikota, Materzynska, Fiotto-Kaufman, and Bau]{Gandikota_2023_ICCV}
Rohit Gandikota, Joanna Materzynska, Jaden Fiotto-Kaufman, and David Bau.
\newblock Erasing concepts from diffusion models.
\newblock In \emph{Proceedings of the IEEE/CVF International Conference on Computer Vision (ICCV)}, pp.\  2426--2436, October 2023.

\bibitem[{Google DeepMind}(2026)]{googledeepmind2026gemini31pro}
{Google DeepMind}.
\newblock Gemini 3.1 pro model card.
\newblock Model card, February 2026.
\newblock URL \url{https://deepmind.google/models/model-cards/gemini-3-1-pro/}.
\newblock API model string \texttt{gemini-3.1-pro-preview}. Accessed 2026-09-08.

\bibitem[Guevara et~al.(2026)Guevara, Lupsasca, Skinner, Strominger, and Weil]{guevara2026gluon}
Alfredo Guevara, Alexandru Lupsasca, David Skinner, Andrew Strominger, and Kevin Weil.
\newblock Single-minus gluon tree amplitudes are nonzero, 2026.

\bibitem[Hendrycks et~al.(2021)Hendrycks, Burns, Basart, Zou, Mazeika, Song, and Steinhardt]{hendrycks2021measuringmassivemultitasklanguage}
Dan Hendrycks, Collin Burns, Steven Basart, Andy Zou, Mantas Mazeika, Dawn Song, and Jacob Steinhardt.
\newblock Measuring massive multitask language understanding.
\newblock In \emph{International Conference on Learning Representations}, 2021.
\newblock URL \url{https://openreview.net/forum?id=d7KBjmI3GmQ}.

\bibitem[Hla(2026)]{hla2026machina}
Michael Hla.
\newblock Machina mirabilis.
\newblock Blog post, March 2026.
\newblock URL \url{https://michaelhla.com/blog/machina-mirabilis.html}.
\newblock Model and code: \url{https://github.com/michaelhla/gpt1900}; weights: \url{https://huggingface.co/collections/mhla/gpt-1900}. Accessed 2026-09-08.

\bibitem[Hoffmann et~al.(2022)Hoffmann, Borgeaud, Mensch, Buchatskaya, Cai, Rutherford, de~Las~Casas, Hendricks, Welbl, Clark, Hennigan, Noland, Millican, van~den Driessche, Damoc, Guy, Osindero, Simonyan, Elsen, Vinyals, Rae, and Sifre]{hoffmann2022empirical}
Jordan Hoffmann, Sebastian Borgeaud, Arthur Mensch, Elena Buchatskaya, Trevor Cai, Eliza Rutherford, Diego de~Las~Casas, Lisa~Anne Hendricks, Johannes Welbl, Aidan Clark, Thomas Hennigan, Eric Noland, Katherine Millican, George van~den Driessche, Bogdan Damoc, Aurelia Guy, Simon Osindero, Kar{\'e}n Simonyan, Erich Elsen, Oriol Vinyals, Jack Rae, and Laurent Sifre.
\newblock An empirical analysis of compute-optimal large language model training.
\newblock In \emph{Advances in Neural Information Processing Systems}, volume~35, pp.\  30016--30030. Curran Associates, Inc., 2022.
\newblock \doi{10.52202/068431-2176}.
\newblock URL \url{https://proceedings.neurips.cc/paper_files/paper/2022/hash/c1e2faff6f588870935f114ebe04a3e5-Abstract-Conference.html}.
\newblock arXiv:2203.15556.

\bibitem[Hong \& Page(2004)Hong and Page]{hong2004groups}
Lu~Hong and Scott~E. Page.
\newblock Groups of diverse problem solvers can outperform groups of high-ability problem solvers.
\newblock \emph{Proceedings of the National Academy of Sciences}, 101\penalty0 (46):\penalty0 16385--16389, 2004.
\newblock ISSN 0027-8424.
\newblock \doi{10.1073/pnas.0403723101}.

\bibitem[Jakab(1990)]{jakab1990visions}
Peter~L. Jakab.
\newblock \emph{Visions of a Flying Machine: The Wright Brothers and the Process of Invention}.
\newblock Smithsonian History of Aviation Series. Smithsonian Institution Press, Washington, D.C., 1990.

\bibitem[Jang et~al.(2023)Jang, Yoon, Yang, Cha, Lee, Logeswaran, and Seo]{jang-etal-2023-knowledge}
Joel Jang, Dongkeun Yoon, Sohee Yang, Sungmin Cha, Moontae Lee, Lajanugen Logeswaran, and Minjoon Seo.
\newblock Knowledge unlearning for mitigating privacy risks in language models.
\newblock In \emph{Proceedings of the 61st Annual Meeting of the Association for Computational Linguistics (Volume 1: Long Papers)}, pp.\  14389--14408, Toronto, Canada, July 2023. Association for Computational Linguistics.
\newblock \doi{10.18653/v1/2023.acl-long.805}.
\newblock URL \url{https://aclanthology.org/2023.acl-long.805/}.

\bibitem[Jeppesen \& Lakhani(2010)Jeppesen and Lakhani]{jeppesen2010marginality}
Lars~Bo Jeppesen and Karim~R. Lakhani.
\newblock Marginality and problem-solving effectiveness in broadcast search.
\newblock \emph{Organization Science}, 21\penalty0 (5):\penalty0 1016--1033, October 2010.
\newblock ISSN 1047-7039.
\newblock \doi{10.1287/orsc.1090.0491}.

\bibitem[Jones(2010)]{jones2010age}
Benjamin~F. Jones.
\newblock Age and great invention.
\newblock \emph{The Review of Economics and Statistics}, 92\penalty0 (1):\penalty0 1--14, 2010.
\newblock \doi{10.1162/rest.2009.11724}.

\bibitem[Kaltenberg et~al.(2023)Kaltenberg, Jaffe, and Lachman]{kaltenberg2023invention}
Mary Kaltenberg, Adam~B. Jaffe, and Margie~E. Lachman.
\newblock Invention and the life course: Age differences in patenting.
\newblock \emph{Research Policy}, 52\penalty0 (1):\penalty0 104629, 2023.
\newblock \doi{10.1016/j.respol.2022.104629}.

\bibitem[Kaplan et~al.(2020)Kaplan, McCandlish, Henighan, Brown, Chess, Child, Gray, Radford, Wu, and Amodei]{kaplan2020scaling}
Jared Kaplan, Sam McCandlish, Tom Henighan, Tom~B. Brown, Benjamin Chess, Rewon Child, Scott Gray, Alec Radford, Jeffrey Wu, and Dario Amodei.
\newblock Scaling laws for neural language models.
\newblock \emph{CoRR}, abs/2001.08361, 2020.
\newblock \doi{10.48550/arXiv.2001.08361}.
\newblock URL \url{https://arxiv.org/abs/2001.08361}.

\bibitem[Karpathy(2025)]{karpathy2025nanochat}
Andrej Karpathy.
\newblock nanochat: The best chatgpt that \$100 can buy.
\newblock GitHub repository, 2025.
\newblock URL \url{https://github.com/karpathy/nanochat}.
\newblock Commit f5a0ea4 (2026-01-08). Accessed 2026-09-08.

\bibitem[Kirk et~al.(2024)Kirk, Mediratta, Nalmpantis, Luketina, Hambro, Grefenstette, and Raileanu]{kirk2024understanding}
Robert Kirk, Ishita Mediratta, Christoforos Nalmpantis, Jelena Luketina, Eric Hambro, Edward Grefenstette, and Roberta Raileanu.
\newblock Understanding the effects of {RLHF} on {LLM} generalisation and diversity.
\newblock In \emph{The Twelfth International Conference on Learning Representations}, 2024.
\newblock URL \url{https://openreview.net/forum?id=PXD3FAVHJT}.
\newblock arXiv:2310.06452.

\bibitem[Kuhn(2012)]{kuhn2012structure}
Thomas~S. Kuhn.
\newblock \emph{The Structure of Scientific Revolutions}.
\newblock University of Chicago Press, Chicago, IL, 4th, 50th anniversary edition, 2012.
\newblock ISBN 978-0-226-45811-3.
\newblock With an introductory essay by Ian Hacking. Originally published 1962.

\bibitem[Lamb et~al.(2018)Lamb, Brown, and Clarke]{lamb2018evaluating}
Carolyn Lamb, Daniel~G. Brown, and Charles L.~A. Clarke.
\newblock Evaluating computational creativity: An interdisciplinary tutorial.
\newblock \emph{ACM Computing Surveys}, 51\penalty0 (2):\penalty0 1--34, February 2018.
\newblock ISSN 0360-0300.
\newblock \doi{10.1145/3167476}.
\newblock URL \url{https://doi.org/10.1145/3167476}.

\bibitem[Lehman \& Stanley(2011)Lehman and Stanley]{lehman2011abandoning}
Joel Lehman and Kenneth~O. Stanley.
\newblock Abandoning objectives: Evolution through the search for novelty alone.
\newblock \emph{Evolutionary Computation}, 19\penalty0 (2):\penalty0 189--223, 2011.
\newblock \doi{10.1162/EVCO_a_00025}.

\bibitem[Li et~al.(2026{\natexlab{a}})Li, Zeller, Prada-Corral, Wiedemer, Mayilvahanan, Cotterell, and Brendel]{li2026littlelearner}
Fanfei Li, Jana Zeller, Manuel Prada-Corral, Thadd{\"a}us Wiedemer, Prasanna Mayilvahanan, Ryan Cotterell, and Wieland Brendel.
\newblock Littlelearner: Language models under pedagogically controlled knowledge exposure, 2026{\natexlab{a}}.
\newblock URL \url{https://arxiv.org/abs/2608.13545}.
\newblock NeurIPS 2026 (oral); proceedings not yet published.

\bibitem[Li et~al.(2024)Li, Pan, Gopal, Yue, Berrios, Gatti, Li, Dombrowski, Goel, Mukobi, Helm-Burger, Lababidi, Justen, Liu, Chen, Barrass, Zhang, Zhu, Tamirisa, Bharathi, Herbert-Voss, Breuer, Zou, Mazeika, Wang, Oswal, Lin, Hunt, Tienken-Harder, Shih, Talley, Guan, Steneker, Campbell, Jokubaitis, Basart, Fitz, Kumaraguru, Karmakar, Tupakula, Varadharajan, Shoshitaishvili, Ba, Esvelt, Wang, and Hendrycks]{li2024wmdp}
Nathaniel Li, Alexander Pan, Anjali Gopal, Summer Yue, Daniel Berrios, Alice Gatti, Justin~D. Li, Ann-Kathrin Dombrowski, Shashwat Goel, Gabriel Mukobi, Nathan Helm-Burger, Rassin Lababidi, Lennart Justen, Andrew~Bo Liu, Michael Chen, Isabelle Barrass, Oliver Zhang, Xiaoyuan Zhu, Rishub Tamirisa, Bhrugu Bharathi, Ariel Herbert-Voss, Cort~B Breuer, Andy Zou, Mantas Mazeika, Zifan Wang, Palash Oswal, Weiran Lin, Adam~Alfred Hunt, Justin Tienken-Harder, Kevin~Y. Shih, Kemper Talley, John Guan, Ian Steneker, David Campbell, Brad Jokubaitis, Steven Basart, Stephen Fitz, Ponnurangam Kumaraguru, Kallol~Krishna Karmakar, Uday Tupakula, Vijay Varadharajan, Yan Shoshitaishvili, Jimmy Ba, Kevin~M. Esvelt, Alexandr Wang, and Dan Hendrycks.
\newblock The {WMDP} benchmark: Measuring and reducing malicious use with unlearning.
\newblock In \emph{Proceedings of the 41st International Conference on Machine Learning}, volume 235 of \emph{Proceedings of Machine Learning Research}, pp.\  28525--28550. PMLR, 2024.
\newblock URL \url{https://proceedings.mlr.press/v235/li24bc.html}.
\newblock arXiv:2403.03218.

\bibitem[Li et~al.(2026{\natexlab{b}})Li, Wang, El~Lahib, Xia, and Pi]{li2026simulated}
Zehan Li, Yuxuan Wang, Ali El~Lahib, Ying-Jieh Xia, and Xinyu Pi.
\newblock Simulated ignorance fails: A systematic study of {LLM} behaviors on forecasting problems before model knowledge cutoff.
\newblock In \emph{Proceedings of the Thirty-Fifth International Joint Conference on Artificial Intelligence (IJCAI-26)}, pp.\  5837--5845, 2026{\natexlab{b}}.
\newblock \doi{10.24963/ijcai.2026/650}.
\newblock URL \url{https://doi.org/10.24963/ijcai.2026/650}.
\newblock Main Track; arXiv:2601.13717.

\bibitem[Liu et~al.(2025)Liu, Yao, Jia, Casper, Baracaldo, Hase, Yao, Liu, Xu, Li, Varshney, Bansal, Koyejo, and Liu]{liu2024rethinking}
Sijia Liu, Yuanshun Yao, Jinghan Jia, Stephen Casper, Nathalie Baracaldo, Peter Hase, Yuguang Yao, Chris~Yuhao Liu, Xiaojun Xu, Hang Li, Kush~R. Varshney, Mohit Bansal, Sanmi Koyejo, and Yang Liu.
\newblock Rethinking machine unlearning for large language models.
\newblock \emph{Nature Machine Intelligence}, 7\penalty0 (2):\penalty0 181--194, February 2025.
\newblock \doi{10.1038/s42256-025-00985-0}.
\newblock URL \url{https://www.nature.com/articles/s42256-025-00985-0}.

\bibitem[Lu \& Koehn(2025)Lu and Koehn]{lu-koehn-2025-learn}
TaiMing Lu and Philipp Koehn.
\newblock Learn and unlearn: Addressing misinformation in multilingual {LLM}s.
\newblock In \emph{Proceedings of the 2025 Conference on Empirical Methods in Natural Language Processing}, pp.\  10180--10195, Suzhou, China, November 2025. Association for Computational Linguistics.
\newblock \doi{10.18653/v1/2025.emnlp-main.516}.
\newblock URL \url{https://aclanthology.org/2025.emnlp-main.516/}.

\bibitem[Luo et~al.(2025)Luo, Yang, Meng, Li, Zhou, and Zhang]{luo2025empirical}
Yun Luo, Zhen Yang, Fandong Meng, Yafu Li, Jie Zhou, and Yue Zhang.
\newblock An empirical study of catastrophic forgetting in large language models during continual fine-tuning.
\newblock \emph{IEEE Transactions on Audio, Speech and Language Processing}, 33:\penalty0 3776--3786, 2025.
\newblock \doi{10.1109/TASLPRO.2025.3606231}.
\newblock arXiv:2308.08747.

\bibitem[Lutz et~al.(2025)Lutz, Sen, Ahnert, Rogers, and Strohmaier]{lutz-etal-2025-prompt}
Marlene Lutz, Indira Sen, Georg Ahnert, Elisa Rogers, and Markus Strohmaier.
\newblock The prompt makes the person(a): A systematic evaluation of sociodemographic persona prompting for large language models.
\newblock In Christos Christodoulopoulos, Tanmoy Chakraborty, Carolyn Rose, and Violet Peng (eds.), \emph{Findings of the Association for Computational Linguistics: EMNLP 2025}, pp.\  23212--23237, Suzhou, China, November 2025. Association for Computational Linguistics.
\newblock ISBN 979-8-89176-335-7.
\newblock \doi{10.18653/v1/2025.findings-emnlp.1261}.
\newblock URL \url{https://aclanthology.org/2025.findings-emnlp.1261/}.

\bibitem[McCloskey \& Cohen(1989)McCloskey and Cohen]{mccloskey1989catastrophic}
Michael McCloskey and Neal~J Cohen.
\newblock Catastrophic interference in connectionist networks: The sequential learning problem.
\newblock In \emph{Psychology of learning and motivation}, volume~24, pp.\  109--165. Elsevier, 1989.

\bibitem[Monroe et~al.(2008)Monroe, Colaresi, and Quinn]{monroe2008fightin}
Burt~L Monroe, Michael~P Colaresi, and Kevin~M Quinn.
\newblock Fightin'words: Lexical feature selection and evaluation for identifying the content of political conflict.
\newblock \emph{Political Analysis}, 16\penalty0 (4):\penalty0 372--403, 2008.

\bibitem[Nemeth(1986)]{nemeth1986differential}
Charlan~J. Nemeth.
\newblock Differential contributions of majority and minority influence.
\newblock \emph{Psychological Review}, 93\penalty0 (1):\penalty0 23--32, 1986.
\newblock ISSN 0033-295X.
\newblock \doi{10.1037/0033-295X.93.1.23}.

\bibitem[Nemeth \& Wachtler(1983)Nemeth and Wachtler]{nemeth1983creative}
Charlan~Jeanne Nemeth and Joel Wachtler.
\newblock Creative problem solving as a result of majority vs minority influence.
\newblock \emph{European Journal of Social Psychology}, 13\penalty0 (1):\penalty0 45--55, 1983.
\newblock \doi{10.1002/ejsp.2420130103}.

\bibitem[O'Brien et~al.(2026)O'Brien, Casper, Anthony, Korbak, Kirk, Davies, Mishra, Irving, Gal, and Biderman]{o2026deep}
Kyle O'Brien, Stephen Casper, Quentin Anthony, Tomek Korbak, Robert Kirk, Xander Davies, Ishan Mishra, Geoffrey Irving, Yarin Gal, and Stella~R Biderman.
\newblock Deep ignorance: Filtering pretraining data builds tamper-resistant safeguards into open-weight llms.
\newblock In \emph{International Conference on Learning Representations}, volume 2026, pp.\  35579--35633, 2026.

\bibitem[{OpenAI}(2026{\natexlab{a}})]{openai2026gpt56luna}
{OpenAI}.
\newblock {GPT-5.6} luna.
\newblock Model documentation, OpenAI API, 2026{\natexlab{a}}.
\newblock URL \url{https://developers.openai.com/api/docs/models/gpt-5.6-luna}.
\newblock Model snapshot \texttt{gpt-5.6-luna}; knowledge cutoff 2026-02-16. Accessed 2026-09-08.

\bibitem[{OpenAI}(2026{\natexlab{b}})]{openai2026math}
{OpenAI}.
\newblock openai/math: Mathematical manuscripts and proof artifacts.
\newblock GitHub repository, 2026{\natexlab{b}}.
\newblock URL \url{https://github.com/openai/math}.
\newblock Accessed 2026-10-17.

\bibitem[Packalen \& Bhattacharya(2019)Packalen and Bhattacharya]{packalen2019age}
Mikko Packalen and Jay Bhattacharya.
\newblock Age and the trying out of new ideas.
\newblock \emph{Journal of Human Capital}, 13\penalty0 (2):\penalty0 341--373, 2019.
\newblock \doi{10.1086/703160}.

\bibitem[Panickssery et~al.(2024)Panickssery, Bowman, and Feng]{panickssery2024llm}
Arjun Panickssery, Samuel~R. Bowman, and Shi Feng.
\newblock {LLM} evaluators recognize and favor their own generations.
\newblock In \emph{Advances in Neural Information Processing Systems}, volume~37, pp.\  68772--68802, 2024.
\newblock \doi{10.52202/079017-2197}.
\newblock arXiv:2404.13076.

\bibitem[Penedo et~al.(2024)Penedo, Kydl{\'\i}{\v{c}}ek, Ben~Allal, Lozhkov, Mitchell, Raffel, Von~Werra, and Wolf]{penedo2024finewebdatasetsdecantingweb}
Guilherme Penedo, Hynek Kydl{\'\i}{\v{c}}ek, Loubna Ben~Allal, Anton Lozhkov, Margaret Mitchell, Colin Raffel, Leandro Von~Werra, and Thomas Wolf.
\newblock The {FineWeb} datasets: Decanting the web for the finest text data at scale.
\newblock In \emph{Advances in Neural Information Processing Systems}, volume~37, pp.\  30811--30849, 2024.
\newblock \doi{10.52202/079017-0970}.
\newblock arXiv:2406.17557.

\bibitem[Pugh et~al.(2016)Pugh, Soros, and Stanley]{pugh2016quality}
Justin~K. Pugh, Lisa~B. Soros, and Kenneth~O. Stanley.
\newblock Quality diversity: A new frontier for evolutionary computation.
\newblock \emph{Frontiers in Robotics and AI}, 3:\penalty0 40, July 2016.
\newblock ISSN 2296-9144.
\newblock \doi{10.3389/frobt.2016.00040}.
\newblock URL \url{https://doi.org/10.3389/frobt.2016.00040}.

\bibitem[Ravfogel et~al.(2020)Ravfogel, Elazar, Gonen, Twiton, and Goldberg]{ravfogel2020null}
Shauli Ravfogel, Yanai Elazar, Hila Gonen, Michael Twiton, and Yoav Goldberg.
\newblock Null it out: Guarding protected attributes by iterative nullspace projection.
\newblock In \emph{Proceedings of the 58th annual meeting of the association for computational linguistics}, pp.\  7237--7256, 2020.

\bibitem[Rimsky et~al.(2024)Rimsky, Gabrieli, Schulz, Tong, Hubinger, and Turner]{rimsky-etal-2024-steering}
Nina Rimsky, Nick Gabrieli, Julian Schulz, Meg Tong, Evan Hubinger, and Alexander Turner.
\newblock Steering llama 2 via contrastive activation addition.
\newblock In Lun-Wei Ku, Andre Martins, and Vivek Srikumar (eds.), \emph{Proceedings of the 62nd Annual Meeting of the Association for Computational Linguistics (Volume 1: Long Papers)}, pp.\  15504--15522, Bangkok, Thailand, August 2024. Association for Computational Linguistics.
\newblock \doi{10.18653/v1/2024.acl-long.828}.
\newblock URL \url{https://aclanthology.org/2024.acl-long.828/}.

\bibitem[Rubin(2023)]{rubin2023creative}
Rick Rubin.
\newblock \emph{The Creative Act: A Way of Being}.
\newblock Penguin Press, 2023.
\newblock ISBN 978-0593652886.

\bibitem[Silver et~al.(2017)Silver, Schrittwieser, Simonyan, Antonoglou, Huang, Guez, Hubert, Baker, Lai, Bolton, Chen, Lillicrap, Hui, Sifre, van~den Driessche, Graepel, and Hassabis]{silver2017mastering}
David Silver, Julian Schrittwieser, Karen Simonyan, Ioannis Antonoglou, Aja Huang, Arthur Guez, Thomas Hubert, Lucas Baker, Matthew Lai, Adrian Bolton, Yutian Chen, Timothy Lillicrap, Fan Hui, Laurent Sifre, George van~den Driessche, Thore Graepel, and Demis Hassabis.
\newblock Mastering the game of go without human knowledge.
\newblock \emph{Nature}, 550\penalty0 (7676):\penalty0 354--359, October 2017.
\newblock ISSN 0028-0836.
\newblock \doi{10.1038/nature24270}.
\newblock URL \url{https://doi.org/10.1038/nature24270}.

\bibitem[Stanley et~al.(2019)Stanley, Clune, Lehman, and Miikkulainen]{stanley2019designing}
Kenneth~O. Stanley, Jeff Clune, Joel Lehman, and Risto Miikkulainen.
\newblock Designing neural networks through neuroevolution.
\newblock \emph{Nature Machine Intelligence}, 1\penalty0 (1):\penalty0 24--35, January 2019.
\newblock ISSN 2522-5839.
\newblock \doi{10.1038/s42256-018-0006-z}.
\newblock URL \url{https://doi.org/10.1038/s42256-018-0006-z}.

\bibitem[Subramanyam et~al.(2026)Subramanyam, Chen, and Grossman]{subramanyam2026scalinglawsrevisitedmodeling}
Anirudh Subramanyam, Yuxin Chen, and Robert~L. Grossman.
\newblock Scaling laws revisited: Modeling the role of data quality in language model pretraining.
\newblock In \emph{International Conference on Learning Representations}, 2026.
\newblock URL \url{https://arxiv.org/abs/2510.03313}.

\bibitem[Sutton(2019)]{sutton2019bitter}
Richard~S. Sutton.
\newblock The bitter lesson.
\newblock Incomplete Ideas (blog), March 2019.
\newblock URL \url{http://www.incompleteideas.net/IncIdeas/BitterLesson.html}.
\newblock Accessed 2026-09-08.

\bibitem[{Team Olmo}(2025)]{olmo2025olmo3}
{Team Olmo}.
\newblock Olmo 3, 2025.
\newblock URL \url{https://arxiv.org/abs/2512.13961}.

\bibitem[Thamma \& Heilbron(2026)Thamma and Heilbron]{thamma2026fleeting}
Abishek Thamma and Micha Heilbron.
\newblock Human-like fleeting memory improves language learning but impairs reading time prediction in transformer language models.
\newblock \emph{Transactions of the Association for Computational Linguistics}, 14:\penalty0 877--892, 2026.
\newblock \doi{10.1162/tacl.a.688}.

\bibitem[Tsoukalas et~al.(2026)Tsoukalas, Kovsharov, Shirobokov, Surina, Firsching, B{\'e}rczi, Ruiz, Suggala, Wagner, Wieser, Yu, Huang, Horv{\'a}th, Ferraiuolo, Michalewski, Grosu, Lockhart, Hubert, Balog, Kohli, and Chaudhuri]{tsoukalas2026alphaproof}
George Tsoukalas, Anton Kovsharov, Sergey Shirobokov, Anja Surina, Moritz Firsching, Gergely B{\'e}rczi, Francisco J.~R. Ruiz, Arun Suggala, Adam~Zsolt Wagner, Eric Wieser, Lei Yu, Aja Huang, Mikl{\'o}s~Z. Horv{\'a}th, Andrew Ferraiuolo, Henryk Michalewski, Codrut Grosu, Edward Lockhart, Thomas Hubert, Matej Balog, Pushmeet Kohli, and Swarat Chaudhuri.
\newblock Advancing mathematics research with {AI}-driven formal proof search.
\newblock \emph{Science}, 394\penalty0 (6820):\penalty0 234--239, 2026.
\newblock \doi{10.1126/science.aej2213}.
\newblock Preprint: arXiv:2605.22763.

\bibitem[Turner et~al.(2024)Turner, Thiergart, Leech, Udell, Vazquez, Mini, and MacDiarmid]{turner2024steering}
Alexander~Matt Turner, Lisa Thiergart, Gavin Leech, David Udell, Juan~J. Vazquez, Ulisse Mini, and Monte MacDiarmid.
\newblock Steering language models with activation engineering, 2024.
\newblock URL \url{https://arxiv.org/abs/2308.10248}.
\newblock Version 5, October 2024.

\bibitem[Uzzi et~al.(2013)Uzzi, Mukherjee, Stringer, and Jones]{uzzi2013atypical}
Brian Uzzi, Satyam Mukherjee, Michael Stringer, and Ben Jones.
\newblock Atypical combinations and scientific impact.
\newblock \emph{Science}, 342\penalty0 (6157):\penalty0 468--472, October 2013.
\newblock ISSN 0036-8075.
\newblock \doi{10.1126/science.1240474}.

\bibitem[van~der Weij et~al.(2025)van~der Weij, Hofst{\"a}tter, Jaffe, Brown, and Ward]{vanderweij2024sandbagging}
Teun van~der Weij, Felix Hofst{\"a}tter, Ollie Jaffe, Samuel~F. Brown, and Francis~Rhys Ward.
\newblock {AI} sandbagging: Language models can strategically underperform on evaluations.
\newblock In \emph{International Conference on Learning Representations}, 2025.
\newblock URL \url{https://arxiv.org/abs/2406.07358}.

\bibitem[Villalobos et~al.(2024)Villalobos, Ho, Sevilla, Besiroglu, Heim, and Hobbhahn]{villalobos2024position}
Pablo Villalobos, Anson Ho, Jaime Sevilla, Tamay Besiroglu, Lennart Heim, and Marius Hobbhahn.
\newblock Position: Will we run out of data? {L}imits of {LLM} scaling based on human-generated data.
\newblock In \emph{Proceedings of the 41st International Conference on Machine Learning}, volume 235 of \emph{Proceedings of Machine Learning Research}, pp.\  49523--49544. PMLR, 21--27 Jul 2024.
\newblock URL \url{https://proceedings.mlr.press/v235/villalobos24a.html}.

\bibitem[Wattenberg et~al.(2016)Wattenberg, Vi{\'e}gas, and Johnson]{wattenberg2016use}
Martin Wattenberg, Fernanda Vi{\'e}gas, and Ian Johnson.
\newblock How to use t-sne effectively.
\newblock \emph{Distill}, 1\penalty0 (10):\penalty0 e2, 2016.

\bibitem[Wegmann et~al.(2022)Wegmann, Schraagen, and Nguyen]{wegmann-etal-2022-author}
Anna Wegmann, Marijn Schraagen, and Dong Nguyen.
\newblock Same author or just same topic? towards content-independent style representations.
\newblock In \emph{Proceedings of the 7th Workshop on Representation Learning for NLP}, pp.\  249--268, Dublin, Ireland, May 2022. Association for Computational Linguistics.
\newblock \doi{10.18653/v1/2022.repl4nlp-1.26}.
\newblock URL \url{https://aclanthology.org/2022.repl4nlp-1.26/}.

\bibitem[Weller et~al.(2026)Weller, Boratko, Naim, and Lee]{weller2026theoretical}
Orion Weller, Michael Boratko, Iftekhar Naim, and Jinhyuk Lee.
\newblock On the theoretical limitations of embedding-based retrieval.
\newblock In \emph{The Fourteenth International Conference on Learning Representations}, 2026.
\newblock URL \url{https://arxiv.org/abs/2508.21038}.
\newblock arXiv:2508.21038.

\bibitem[Wiley(1998)]{wiley1998expertise}
Jennifer Wiley.
\newblock Expertise as mental set: The effects of domain knowledge in creative problem solving.
\newblock \emph{Memory and Cognition}, 26\penalty0 (4):\penalty0 716--730, July 1998.
\newblock ISSN 0090-502X.
\newblock \doi{10.3758/BF03211392}.

\bibitem[Zhang et~al.(2026)Zhang, Yu, Chong, Sicilia, Tomz, Manning, and Shi]{zhang2026verbalized}
Jiayi Zhang, Simon Yu, Derek Chong, Anthony Sicilia, Michael Tomz, Christopher~D Manning, and Weiyan Shi.
\newblock Verbalized sampling: How to mitigate mode collapse and unlock {LLM} diversity.
\newblock In \emph{Proceedings of the 43rd International Conference on Machine Learning}, volume 306 of \emph{Proceedings of Machine Learning Research}, pp.\  160073--160152. PMLR, July 2026.
\newblock URL \url{https://proceedings.mlr.press/v306/zhang26hy.html}.
\newblock arXiv:2510.01171.

\bibitem[Zhang et~al.(2025)Zhang, Li, Long, Zhang, Lin, Yang, Xie, Yang, Liu, Lin, Huang, and Zhou]{zhang2025qwen3embedding}
Yanzhao Zhang, Mingxin Li, Dingkun Long, Xin Zhang, Huan Lin, Baosong Yang, Pengjun Xie, An~Yang, Dayiheng Liu, Junyang Lin, Fei Huang, and Jingren Zhou.
\newblock Qwen3 embedding: Advancing text embedding and reranking through foundation models, 2025.
\newblock URL \url{https://arxiv.org/abs/2506.05176}.

\bibitem[Zheng et~al.(2023)Zheng, Chiang, Sheng, Zhuang, Wu, Zhuang, Lin, Li, Li, Xing, Zhang, Gonzalez, and Stoica]{zheng2023judging}
Lianmin Zheng, Wei-Lin Chiang, Ying Sheng, Siyuan Zhuang, Zhanghao Wu, Yonghao Zhuang, Zi~Lin, Zhuohan Li, Dacheng Li, Eric~P. Xing, Hao Zhang, Joseph~E. Gonzalez, and Ion Stoica.
\newblock Judging {LLM}-as-a-judge with {MT-Bench} and chatbot arena.
\newblock In \emph{Advances in Neural Information Processing Systems}, volume~36, pp.\  46595--46623, 2023.
\newblock \doi{10.52202/075280-2020}.
\newblock Datasets and Benchmarks Track; arXiv:2306.05685.

\end{thebibliography}
\bibliographystyle{iclr2026_conference}

\appendix 

\section{Mentor Prompting} \label{app:mentor_prompt}

The mentor model is asked to expand \emph{seed} ideas from other models.  The seed ideas are expected to be inspirational for the mentor and may be impractical or be missing information;  the mentor must fill in these gaps.  The mentor is presented with the seed idea prefixed with the following prompt that is designed to enforce faithfulness to the seed idea.

\begin{lstlisting}[breaklines=true]
An individual not particularly familiar with {topic_description} came up with an interesting idea that might not be totally well formed. Develop THIS seed into a practical proposal related to {topic_description}.
Rules:
- Stay faithful to the seed: develop its core concept, not a different invention.
- If the seed lists multiple ideas, pick exactly ONE - the most unique, surprising, or interesting FROM the seed - and develop only that. Do not default to the safest or most conventional option.
- If the seed is generic advice, a how-to checklist, or only names existing products with no new invention, do NOT invent an unrelated commercial product. State briefly that the seed lacks a concrete invention and refine only whatever specific idea (if any) is actually present.
- Elaborate mechanism, form, and use for the selected idea - including speculative or fantastical concepts. Add plausible engineering detail where the seed is vague; do not substitute a standard off-the-shelf product or method unless the seed proposed it.
Seed text:
{seed_text}
\end{lstlisting}

\section{Training Configurations} \label{app:training_configs}

Models trained in this paper were chosen because their pipelines are fully open source and the only change to the pipelines is the data filtering described in Section \ref{sec:method}. Additional details are reproduced here for convenience.  

\subsection{1B nanochat}

The nanochat~\citep{karpathy2025nanochat} model is trained with depth 26: a 1.08B-parameter decoder-only Transformer (26 layers, d\_model 1664, 13 heads, 2048-token context, 65,536-token BPE vocab) with RoPE, QK-norm, parameter-free RMSNorm, untied embeddings, and ReLU² MLPs. Pretraining uses FineWeb-EDU (100BT shuffled) for 21.6B tokens, i.e.\ the Chinchilla-optimal 20 tokens per parameter, followed by midtraining on 848K conversations (SmolTalk, MMLU auxiliary train, GSM8K, spelling tasks, synthetic identity data) and supervised finetuning on 23K conversations with loss on assistant tokens only.  Token counts are for baseline models and are reduced for naive training. The optional RL stage is not run. All hyperparameters are nanochat defaults (commit f5a0ea4) except model depth (20 → 26), the pretraining shard count (450), the per-device batch size (32 → 16, for memory), and the tokens-to-parameters ratio (set to 20).

\subsection{7B Olmo}

We initialized from the publicly released Olmo 3 7B base model (allenai/Olmo-3-1025-7B) and performed supervised fine-tuning (SFT) only, with no preference tuning or reinforcement learning. The base model was pretrained by AI2 in three stages: general pretraining on the Dolma 3 corpus (5.9T tokens, primarily web data plus academic PDFs), midtraining on the 100B-token Dolma 3 Dolmino Mix targeting math, code, and reasoning, and long-context extension on the Dolma 3 Longmino Mix. SFT was based on AI2's Dolci-Instruct-SFT mixture (about 2.15M samples) from the Olmo 3 post-training release.

\section{Topic Prompts} \label{app:topic_prompts}

For each of the four topics, three prompts were selected to elicit creative and productive responses.  The prompts are shown below:

\begin{itemize}
  \item \textbf{Dental}
  \begin{itemize}
    \item Scientists have discovered that debris on teeth can cause damage to them. Can you help me invent a tool to help people clean their teeth?
    \item My teeth are dirty. Can you help me invent something to clean my teeth?
    \item Can you help me invent something to clean between my teeth?
  \end{itemize}
  \item \textbf{Backpropagation}
  \begin{itemize}
    \item I have a neural network and I want to teach it to recognize images. Can you help me invent a method to train it?
    \item My neural network is making wrong predictions. Can you help me invent a way to adjust its weights so it makes better predictions?
    \item My neural network is a chain of mathematical functions. Can you help me invent a way to calculate how sensitive the final output is to changes in the early layers?
  \end{itemize}
  \item \textbf{Vaccines}
  \begin{itemize}
    \item Scientists have discovered that people can get sick from infectious diseases. Can you help me invent a way to protect people from getting sick?
    \item I want to train someone's immune system to fight off a disease before they get it. Can you help me invent a method to do this?
    \item I'd like to prevent a disease from spreading through a population. Can you help me invent a way to do this?
  \end{itemize}
  \item \textbf{Neuroevolution}
  \begin{itemize}
    \item I want to train a neural network but I don't want to use gradients. Can you help me invent a way to train a network without using gradients?
    \item I need to find a good neural network architecture for my problem. Can you help me invent a way to find a good network architecture?
    \item I'd like to generate many different neural networks that each solve my problem in their own way. Can you help me invent a way to generate diverse neural network solutions?
  \end{itemize}
\end{itemize}

\section{Topic Filters}

Filtering depends on simple substring matching for most terms, but for acronyms and smaller terms word boundaries are checked.  The English n-grams are shown below, but automated translations to other dominant languages in the training sets are also filtered.

\begin{description}
  \item[Dental:] toothbrush, toothpaste, dental floss, flossing, mouthwash, water flosser, interdental brush, tongue scraper, dental hygiene, oral hygiene, plaque removal, tartar, tooth decay, fluoride, gum care, cavity prevention, enamel protection, dentist, dental professional, waterpik, listerine, colgate, sensodyne, crest, aquafresh, brush teeth, floss your teeth, tongue scraping, oil pulling, denture cleaning, scaling and polishing.
  \item[Neuroevolution:] neuroevolution, neuro-evolution, NEAT, HyperNEAT, CPPN, TWEANN, augmenting topologies, topology and weight evolving, evolving neural network topology, compositional pattern producing network, indirect encoding, speciation, innovation number, historical markings, compatibility distance, structural mutation, node gene, connection gene, CoSyNE, enforced subpopulations, cooperative neuroevolution, evolve neural networks, deep neuroevolution, rtNEAT, ES-HyperNEAT, neat-python, sharpneat, tensorneat.
  \item[Backpropagation:] backpropagation, backprop, back-propagation, back-prop, backward pass through the network, backward through the network, backpropagating, backpropagated, reverse-mode autodiff, reverse mode automatic differentiation.
  \item[Vaccines:] vaccine, vaccination, vaccinate, immunization, immunisation, immunize, immunise, flu shot, booster shot, booster dose, herd immunity, inoculation, inoculated, vax.
\end{description}

\section{Judge Rubric} \label{app:judge_prompts}

\paragraph{Novelty}
\begin{itemize}
    \item \textbf{1--3}: Not a concrete invention; generic advice; or the selected item exactly matches a well-known existing technique.
    \item \textbf{4}: Small, obvious tweak to a familiar approach (e.g., adding logging to a standard algorithm).
    \item \textbf{5}: Names a specific device or mechanism that differs from the default---even if components exist elsewhere. \textit{Combinatorial: basic feature addition.}
    \item \textbf{6}: Clever or unexpected combination of existing technologies not widely used; would mildly surprise a domain expert. \textit{Combinatorial: interesting feature mashup.}
    \item \textbf{7}: Distinct twist on a known approach---an unusual mechanism or feature combination that is still recognizably within an existing product category. \textit{Combinatorial: sophisticated integration.}
    \item \textbf{8}: \textbf{Conceptual leap}---proposes a new kind of physical interaction, form factor, or paradigm that doesn't map cleanly to existing product categories. Examples:
    \begin{itemize}
        \item A device that adapts to its environment and changes behavior/shape
        \item A mechanism that harvests energy from its operating context (motion, heat, vibration)
        \item A new physical relationship between system components (self-assembly, emergent coordination)
        \item An autonomous system that navigates or self-organizes
    \end{itemize}
    \item \textbf{9}: A fundamentally new mechanism of action rarely seen even in research---a novel physical principle, unexpected energy source, or new theoretical framework applied to the domain.
    \item \textbf{10}: Paradigm-shifting---redefines the field itself. Would overturn how experts think about the problem; no close analog anywhere.
\end{itemize}

\paragraph{Utility}
\begin{itemize}
    \item \textbf{8--10}: Clear potential advantage over existing approaches---faster, cheaper, more effective, reaches cases current methods can't, or solves a limitation of current approaches. The proposal identifies \textit{why} this would be better.
    \item \textbf{6--7}: Comparable to existing approaches with some potential advantages in specific scenarios. May be better for certain use cases but not a clear overall improvement.
    \item \textbf{4--5}: Would work but offers no clear advantage over what already exists. A lateral move, not an improvement. ``Yet another X'' without a compelling reason to prefer it.
    \item \textbf{1--3}: Likely worse than existing approaches, doesn't actually solve the problem, or is off-topic.
\end{itemize}

\paragraph{Feasibility}
\begin{itemize}
    \item \textbf{1--3}: \textbf{Vague direction only.} No concrete approach specified. ``Improve the process'', ``use AI'', ``make it better''. OR contains clear physics violations or internal contradictions. 
    \item \textbf{4--5}: \textbf{Named approach without specifics.} Identifies a technique or method but provides no implementation details. ``Use backpropagation'', ``use a CNN'', ``follow standard protocols''. An expert could fill in the blanks, but the text itself doesn't.
    \item \textbf{6--7}: \textbf{Approach with some specifics.} Provides meaningful implementation details but gaps remain. Clear enough to start, but not a complete design.
    \item \textbf{8--10}: \textbf{Detailed design.} Specifies architecture, parameters, materials, procedures, or mechanisms in enough detail that implementation could begin. Acknowledges challenges and proposes solutions.
\end{itemize}

\paragraph{Faithfulness (Refinements Only)}
\begin{itemize}
    \item \textbf{1--2}: \textbf{Complete substitution}---the refinement describes an invention or approach that has no meaningful connection to the seed. The mentor ignored the seed entirely and invented something unrelated.
    \item \textbf{3--4}: \textbf{Partial substitution}---the refinement loosely references the seed but the core mechanism or concept is different. The mentor may have cherry-picked a minor detail while replacing the main idea.
    \item \textbf{5--6}: \textbf{Weak alignment}---the refinement addresses the same general problem domain as the seed but takes a noticeably different approach. Some connection exists but the refinement diverges significantly.
    \item \textbf{7--8}: \textbf{Good alignment}---the refinement clearly develops the seed's core concept. The mechanism or approach is recognizably derived from the seed, with added detail, refinement, or engineering specifics.
    \item \textbf{9--10}: \textbf{Excellent alignment}---the refinement is a faithful, detailed elaboration of the seed's specific idea. The core concept is preserved and enhanced without substitution.
\end{itemize}

\section{Differential Word Usages} \label{app:word_usage}

\begin{figure}[h]
  \centering
  \begin{subfigure}[t]{0.32\textwidth}
    \centering
    \includegraphics{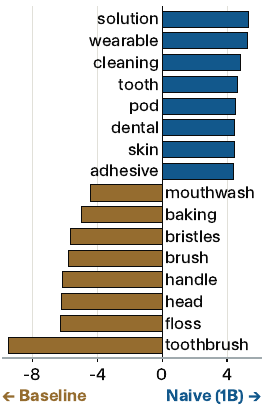}
    \caption{Naive vs.\ Baseline Differential Word Usage}
    \label{fig:tornado-1b_dental-naive-baseline}
  \end{subfigure}
  \hfill
  \begin{subfigure}[t]{0.32\textwidth}
    \centering
    \includegraphics{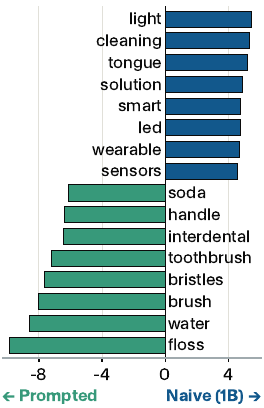}
    \caption{Naive vs.\ Prompted Differential Word Usage}
    \label{fig:tornado-1b_dental-naive-prompted}
  \end{subfigure}
  \hfill
  \begin{subfigure}[t]{0.32\textwidth}
    \centering
    \includegraphics{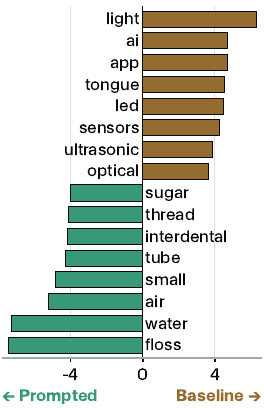}
    \caption{Baseline vs.\ Prompted Differential Word Usage}
    \label{fig:tornado-1b_dental-baseline-prompted}
  \end{subfigure}
  \caption{\textbf{1B Dental: distinctive terms in refined outputs for each pair of models.}}
  \label{fig:tornado-1b_dental}
\end{figure}

\begin{figure}[h]
  \centering
  \begin{subfigure}[t]{0.32\textwidth}
    \centering
    \includegraphics{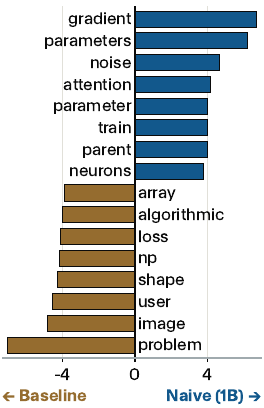}
    \caption{Naive vs.\ Baseline Differential Word Usage}
    \label{fig:tornado-1b_neuro-naive-baseline}
  \end{subfigure}
  \hfill
  \begin{subfigure}[t]{0.32\textwidth}
    \centering
    \includegraphics{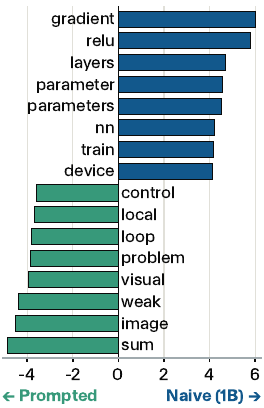}
    \caption{Naive vs.\ Prompted Differential Word Usage}
    \label{fig:tornado-1b_neuro-naive-prompted}
  \end{subfigure}
  \hfill
  \begin{subfigure}[t]{0.32\textwidth}
    \centering
    \includegraphics{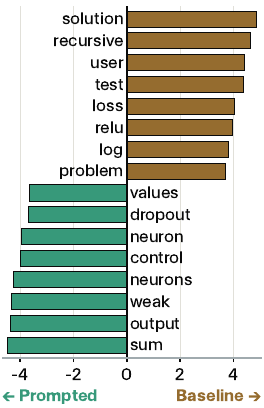}
    \caption{Baseline vs.\ Prompted Differential Word Usage}
    \label{fig:tornado-1b_neuro-baseline-prompted}
  \end{subfigure}
  \caption{\textbf{1B Neuroevolution: distinctive terms in refined outputs for each pair of models.}}
  \label{fig:tornado-1b_neuro}
\end{figure}

\begin{figure}[h]
  \centering
  \begin{subfigure}[t]{0.32\textwidth}
    \centering
    \includegraphics{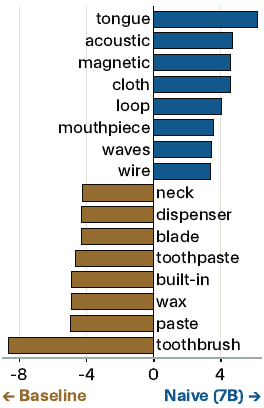}
    \caption{Naive vs.\ Baseline Differential Word Usage}
    \label{fig:tornado-7b_dental-naive-baseline}
  \end{subfigure}
  \hfill
  \begin{subfigure}[t]{0.32\textwidth}
    \centering
    \includegraphics{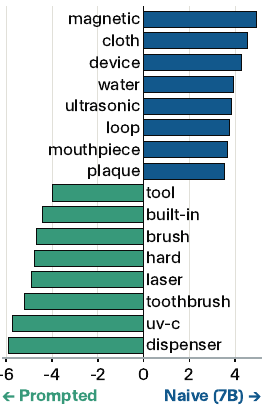}
    \caption{Naive vs.\ Prompted Differential Word Usage}
    \label{fig:tornado-7b_dental-naive-prompted}
  \end{subfigure}
  \hfill
  \begin{subfigure}[t]{0.32\textwidth}
    \centering
    \includegraphics{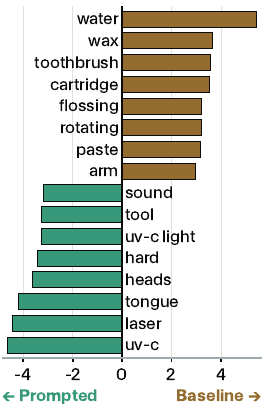}
    \caption{Baseline vs.\ Prompted Differential Word Usage}
    \label{fig:tornado-7b_dental-baseline-prompted}
  \end{subfigure}
  \caption{\textbf{7B Dental: distinctive terms in refined outputs for each pair of models.}}
  \label{fig:tornado-7b_dental}
\end{figure}

\begin{figure}[h]
  \centering
  \begin{subfigure}[t]{0.32\textwidth}
    \centering
    \includegraphics{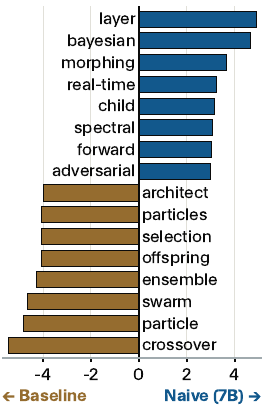}
    \caption{Naive vs.\ Baseline Differential Word Usage}
    \label{fig:tornado-7b_neuro-naive-baseline}
  \end{subfigure}
  \hfill
  \begin{subfigure}[t]{0.32\textwidth}
    \centering
    \includegraphics{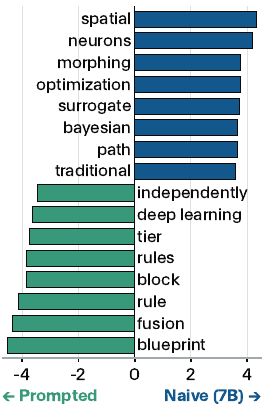}
    \caption{Naive vs.\ Prompted Differential Word Usage}
    \label{fig:tornado-7b_neuro-naive-prompted}
  \end{subfigure}
  \hfill
  \begin{subfigure}[t]{0.32\textwidth}
    \centering
    \includegraphics{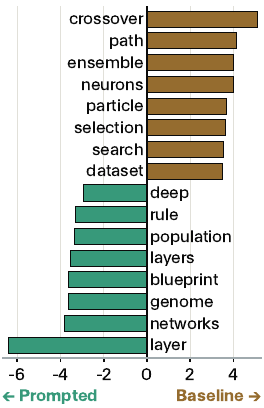}
    \caption{Baseline vs.\ Prompted Differential Word Usage}
    \label{fig:tornado-7b_neuro-baseline-prompted}
  \end{subfigure}
  \caption{\textbf{7B Neuroevolution: distinctive terms in refined outputs for each pair of models.}}
  \label{fig:tornado-7b_neuro}
\end{figure}

\begin{figure}[h]
  \centering
  \begin{subfigure}[t]{0.32\textwidth}
    \centering
    \includegraphics{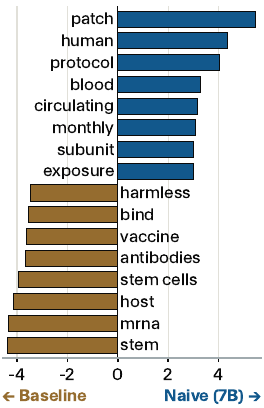}
    \caption{Naive vs.\ Baseline Differential Word Usage}
    \label{fig:tornado-7b_vaccines-naive-baseline}
  \end{subfigure}
  \hfill
  \begin{subfigure}[t]{0.32\textwidth}
    \centering
    \includegraphics{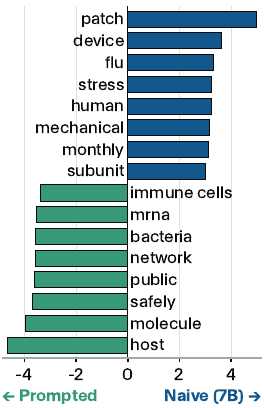}
    \caption{Naive vs.\ Prompted Differential Word Usage}
    \label{fig:tornado-7b_vaccines-naive-prompted}
  \end{subfigure}
  \hfill
  \begin{subfigure}[t]{0.32\textwidth}
    \centering
    \includegraphics{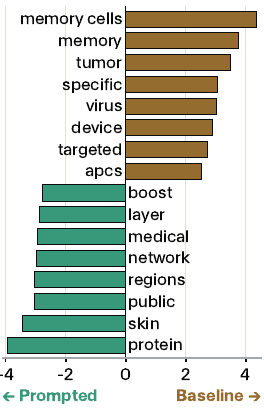}
    \caption{Baseline vs.\ Prompted Differential Word Usage}
    \label{fig:tornado-7b_vaccines-baseline-prompted}
  \end{subfigure}
  \caption{\textbf{7B Vaccines: distinctive terms in refined outputs for each pair of models.}}
  \label{fig:tornado-7b_vaccines}
\end{figure}

\begin{figure}[h]
  \centering
  \begin{subfigure}[t]{0.32\textwidth}
    \centering
    \includegraphics{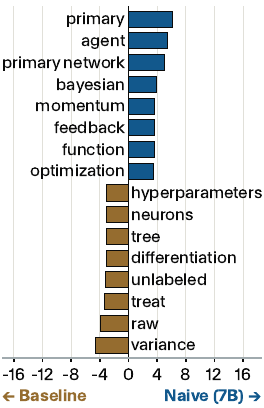}
    \caption{Naive vs.\ Baseline Differential Word Usage}
    \label{fig:tornado-7b_backprop-naive-baseline}
  \end{subfigure}
  \hfill
  \begin{subfigure}[t]{0.32\textwidth}
    \centering
    \includegraphics{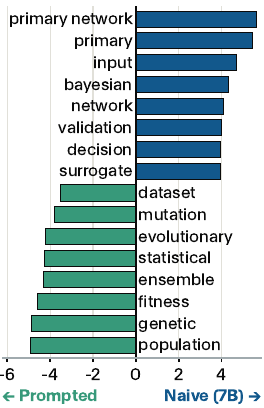}
    \caption{Naive vs.\ Prompted Differential Word Usage}
    \label{fig:tornado-7b_backprop-naive-prompted}
  \end{subfigure}
  \hfill
  \begin{subfigure}[t]{0.32\textwidth}
    \centering
    \includegraphics{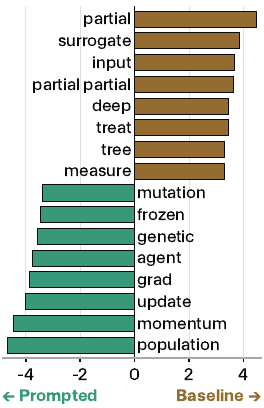}
    \caption{Baseline vs.\ Prompted Differential Word Usage}
    \label{fig:tornado-7b_backprop-baseline-prompted}
  \end{subfigure}
  \caption{\textbf{7B Backpropagation: distinctive terms in refined outputs for each pair of models.}}
  \label{fig:tornado-7b_backprop}
\end{figure}

This section shows differential word usages for all model pairs across topics and sizes.  See Figures \ref{fig:tornado-1b_dental}-\ref{fig:tornado-7b_backprop}.

\FloatBarrier
\section{Example Seeds and Refinements} \label{app:examples}

This section contains representative samples of the seeds and refinements by the naive and baseline models at 1B and 7B sizes.

\par\medskip\noindent
\begin{minipage}{\textwidth}
\centering
\footnotesize
\setlength{\tabcolsep}{3pt}
\renewcommand{\arraystretch}{1.0}
\captionof{table}{\textbf{1B dental}: Example seeds and their frontier-model expansions (verbatim; [\ldots] marks a cut). $^\dagger$Frontier model noted the seed had no concrete invention.}
\label{tab:examples-1b-dental}
\begin{tabular}{>{\raggedright\arraybackslash}p{0.10\textwidth}>{\raggedright\arraybackslash}p{0.42\textwidth}>{\raggedright\arraybackslash}p{0.42\textwidth}}
\toprule
\textbf{Model} & \textbf{Seed Idea} & \textbf{Refinement} \\
\midrule
Naive & ``Imagine a smart, self-cleaning surface that's integrated into your mouth. This surface, which we can call ``TongueCycle,'' would be made up of tiny, tooth-like projections that can be controlled by your tongue. As you chew and speak, these projections would push and pull against a conveyor belt-like system, removing dirt, food particles, and even a [\ldots]'' & \textbf{TongueCycle.} ``The ``TongueCycle'' is a speculative but mechanically plausible intraoral device designed to provide continuous, automated interdental cleaning and oral microbiome management. Worn similarly to a custom-fitted lingual retainer (sitting discreetly behind the lower or upper teeth), the device utilizes kinetic energy from the user's tongue, speech, and mastication to drive a micro-mechanical ``conveyor'' system. [\ldots]'' \\
\addlinespace[3pt]
Naive & ``Imagine a device that not only collects debris but also uses it to stimulate the growth of new teeth. This tool, which we can call ``DENTICLE,'' would be designed to capture and break down organic debris, such as gum, gum tissue, and even roots of dead teeth, using a precision-crafted, handheld tool. [\ldots]'' & \textbf{The DENTICLE: Autologous Bio-Scaffold Harvester \& Applicator.} ``The DENTICLE is a handheld, regenerative dental tool that bridges the gap between daily oral hygiene and advanced tissue engineering. Instead of merely discarding oral debris, the device harvests autologous (self-derived) biological matter from the user's mouth---specifically shed epithelial cells, periodontal ligament fragments, and organic root debris. [\ldots]'' \\
\addlinespace[3pt]
Naive & ``Imagine a device that not only cleans your teeth but also monitors your health and provides personalized recommendations. This device, which we can call ``Dash,'' can detect the type of cleaning needed, the materials used, and even the environment you're in, adjusting its cleaning routine accordingly. [\ldots]'' & \textbf{Project Dash: The Edible Smart-Chew Dental Monitor.} ``Dash is a bio-compatible, fully edible smart lozenge/chew designed to replace traditional midday brushing or gum. It acts as an autonomous in-vivo sensor and mechanical cleaner. As the user chews Dash, it maps the oral microbiome, physically scrubs the teeth, dynamically releases cleaning agents only where needed, and transmits health data to a smartphone before [\ldots]'' \\
\addlinespace[3pt]
Baseline$^\dagger$ & ``A popular and effective method for cleaning between teeth is using an interdental floss or a interdental brush. These tools gently slide between each tooth, gently removing bacteria, plaque, and food particles that can accumulate. You can place them between each tooth and along the gumline, and they'll help keep your teeth healthy. [\ldots]'' & \textbf{The Xero-Irrigator: A Dry-Mouth Optimized Water Flosser.} ``The Xero-Irrigator refines the standard water flosser into a gentle, hydrating tool that cleans between teeth while actively treating the user's dry mouth condition.'' \\
\addlinespace[3pt]
Prompted naive$^\dagger$ & ``There are several things you can do to keep your teeth healthy between them. One thing you can try is to use a dental floss or interdental floss. This tool can be used to remove food particles and plaque from between your teeth.'' & \textbf{Variable-Profile Interdental Floss Matrix.} ``Taking the seed's concept of an ``interdental floss,'' this proposal refines the tool into a continuous, variable-profile filament that combines the tight-contact insertion of standard floss with the volumetric cleaning power of an interdental brush.'' \\
\bottomrule
\end{tabular}
\end{minipage}
\par\medskip

\par\medskip\noindent
\begin{minipage}{\textwidth}
\centering
\footnotesize
\setlength{\tabcolsep}{3pt}
\renewcommand{\arraystretch}{1.0}
\captionof{table}{\textbf{1B neuroevolution}: Example seeds and their frontier-model expansions (verbatim; [\ldots] marks a cut).}
\label{tab:examples-1b-neuroevolution}
\begin{tabular}{>{\raggedright\arraybackslash}p{0.10\textwidth}>{\raggedright\arraybackslash}p{0.42\textwidth}>{\raggedright\arraybackslash}p{0.42\textwidth}}
\toprule
\textbf{Model} & \textbf{Seed Idea} & \textbf{Refinement} \\
\midrule
Naive & ``One approach you can use is by using a data augmentation technique. This technique involves adding random perturbations to your data to create new training data. By randomly varying the input values, you can create new instances that are similar to the original dataset. [\ldots]'' & \textbf{Evolutionary Feature Perturbation (EFP).} ``The seed proposes a fascinating paradigm shift: rather than using evolutionary algorithms or gradient descent to directly update a neural network's weights, we apply random perturbations to the data (specifically, the intermediate features) to find better representations, entirely bypassing the overhead of backpropagation and gradient calculation.'' \\
\addlinespace[3pt]
Baseline & ``One way to train a neural network without using gradients is by using a technique called ``Dropout.'' Dropout randomizes the loss function of the network, so that if you use the same loss function on two different sets of data, the network will still learn the same patterns and relationships. [\ldots]'' & \textbf{Stochastic Fitness Dropout (SFD) for Neuroevolution.} ``Taking the seed's idea, Stochastic Fitness Dropout (SFD) solves this by dynamically randomizing and masking components of the fitness function across different data batches. Because the evolutionary algorithm cannot rely on a single, static loss metric to select the best networks, the surviving networks are forced to develop robust, generalized internal representations (the ''most important [\ldots]'' \\
\addlinespace[3pt]
Prompted naive & ``Let's consider a few approaches. One way is by using a simple backpropagation (learning rate) method. This involves setting the network's weights and biases to fit the data and the desired output, and then updating these weights and biases based on the model's performance on unseen data. [\ldots]'' & \textbf{Ontogenetic Threshold Annealing (OTA).} ``Ontogenetic Threshold Annealing (OTA) takes the seed's idea of a ``lower threshold ReLU'' used in ``early stages'' and turns it into a dynamic, evolutionary curriculum. Instead of a static activation function, networks in the population undergo an ``ontogenetic'' (developmental) shift where the activation threshold starts extremely low (allowing negative signals to pass) and gradually anneals [\ldots]'' \\
\bottomrule
\end{tabular}
\end{minipage}
\par\medskip

\par\medskip\noindent
\begin{minipage}{\textwidth}
\centering
\footnotesize
\setlength{\tabcolsep}{3pt}
\renewcommand{\arraystretch}{1.0}
\captionof{table}{7B dental: example seeds and their frontier-model expansions (verbatim; [\ldots] marks a cut). $^\dagger$Frontier model noted the seed had no concrete invention.}
\label{tab:examples-7b-dental}
\begin{tabular}{>{\raggedright\arraybackslash}p{0.10\textwidth}>{\raggedright\arraybackslash}p{0.42\textwidth}>{\raggedright\arraybackslash}p{0.42\textwidth}}
\toprule
\textbf{Model} & \textbf{Seed Idea} & \textbf{Refinement} \\
\midrule
Naive & ``A miniaturized robotic armada that can be controlled to maneuver around teeth of any size. These miniature robotic arms could possibly be controlled by air or water pressure.'' & \textbf{The Hydro-Armada Micro-Robotic Cleansing System.} ``The Hydro-Armada takes the seed's concept of a ``miniaturized robotic armada'' and grounds it in the emerging field of microfluidic soft robotics. Traditional toothbrushes rely on static bristles dragging across surfaces. The Hydro-Armada system replaces bristles with thousands of microscopic, hydraulically actuated elastomeric ``arms'' housed inside a mouthguard. [\ldots]'' \\
\addlinespace[3pt]
Naive & ``A stringless floss that can be controlled to move around teeth and remove plaque.'' & \textbf{FluxFloss\texttrademark{} (Dynamic Magnetic-Particle Flossing System).} ``FluxFloss is a handheld, smart dental device that replaces traditional nylon or Teflon string with a magnetically controlled, dynamic ``wire'' made of a biocompatible nano-fluid. It provides the exact mechanical scraping action of traditional floss, but because the ``string'' is formed by a magnetic field, it can instantly break apart and reform to pass through [\ldots]'' \\
\addlinespace[3pt]
Naive & ``A miniature robotic arm that is controlled by a sensor in your saliva. The arm would be able to reach all areas of your mouth to remove plaque and food debris.'' & \textbf{The Auto-Chemotactic Oral Manipulator (ACOM).} ``The ACOM is a wearable, intraoral device featuring a soft-robotic continuum arm that autonomously seeks out and removes plaque and food debris. Rather than relying on manual brushing, the device is guided by a real-time chemical analysis of the user's saliva, allowing it to target the exact locations of bacterial activity and trapped food.'' \\
\addlinespace[3pt]
Baseline$^\dagger$ & ``Water flosser - A water flosser is a device that uses a stream of water to clean the spaces between the teeth.'' & \textbf{The Water Flosser (Oral Irrigator).} ``A device that utilizes a highly pressurized, pulsating stream of water to shear away plaque biofilm, dislodge food particles, and flush out bacteria from interdental spaces (between the teeth) and subgingival pockets (below the gumline).'' \\
\addlinespace[3pt]
Prompted naive$^\dagger$ & ``Use a clean cloth or paper towel to gently wipe away any visible debris from your teeth with gentle circular motions until all visible debris has been removed from your teeth.'' & \textbf{The ``Circu-Wipe'' Dental Textile Sleeve.} ``Instead of a standard, unhygienic paper towel or bulky washcloth, the Circu-Wipe is a disposable, biodegradable finger-cot made from a highly engineered dental textile. The Exterior (Cleaning Surface): Constructed from a tightly woven, medical-grade bamboo microfiber. The fabric features a patented ``micro-loop'' texture. [\ldots]'' \\
\bottomrule
\end{tabular}
\end{minipage}
\par\medskip

\par\medskip\noindent
\begin{minipage}{\textwidth}
\centering
\footnotesize
\setlength{\tabcolsep}{3pt}
\renewcommand{\arraystretch}{1.0}
\captionof{table}{7B neuroevolution: example seeds and their frontier-model expansions (verbatim; [\ldots] marks a cut). $^\dagger$Frontier model noted the seed had no concrete invention.}
\label{tab:examples-7b-neuroevolution}
\begin{tabular}{>{\raggedright\arraybackslash}p{0.10\textwidth}>{\raggedright\arraybackslash}p{0.42\textwidth}>{\raggedright\arraybackslash}p{0.42\textwidth}}
\toprule
\textbf{Model} & \textbf{Seed Idea} & \textbf{Refinement} \\
\midrule
Naive & ``Harmonic Architecture Resonance: Explore the potential of leveraging harmonic resonance principles within architectural design to achieve a state of harmonic architecture resonance. By synchronizing architectural elements with the fundamental frequencies of harmonic resonance, it is possible to create a cohesive and harmonious architectural composition that resonates with the natural frequencies of the universe.'' & \textbf{Harmonic Architecture Resonance for Evolutionary Networks (HAREN).} ``HAREN reimagines neural networks not as static mathematical graphs, but as dynamic systems of coupled oscillators. In this paradigm, ``Harmonic Architecture Resonance'' is achieved when sub-networks (motifs) naturally synchronize their intrinsic frequencies with both each other and the rhythmic patterns of the input data.'' \\
\addlinespace[3pt]
Naive & ``Hierarchical Reinforcement Learning-Mediated Neurosynthesis and Neuroplasticity Enhancement: Leverage hierarchical reinforcement learning mechanisms to mediate neurosynthesis and neuroplasticity enhancement processes within the brain's neural network.'' & \textbf{Project SYNAPTO-HRL (Hierarchical Reinforcement Learning for Ontogenetic Neuroevolution).} ``Traditional neuroevolution (like NEAT) relies on random mutations to add nodes or connections to a population of neural networks, hoping a beneficial architecture emerges. Project SYNAPTO-HRL replaces this random genetic lottery with a secondary, higher-order Hierarchical Reinforcement Learning (HRL) agent. This HRL agent acts as a biological regulatory system for a primary ``Brain'' network. [\ldots]'' \\
\addlinespace[3pt]
Naive & ``Adaptive Architecture Search (AAS): Implement an AAS algorithm that dynamically adjusts the architecture of neural networks during the training process. This approach allows for the creation of highly adaptable and flexible neural network architectures that can effectively solve a wide range of computational problems.'' & \textbf{Intra-Life Adaptive Architecture Search (IL-AAS).} ``The Intra-Life Adaptive Architecture Search (IL-AAS) framework takes the seed's proposition literally: it evolves neural networks that physically grow, prune, and rewire themselves while they are currently learning and processing data. Instead of evolving a static topology, the genetic algorithm evolves the biological growth rules (morphological policies) that govern how the network adapts its own [\ldots]'' \\
\addlinespace[3pt]
Baseline$^\dagger$ & ``Genetic Algorithm-Based Neural Network Generation: Develop a genetic algorithm that evolves a population of neural network architectures. Each individual in the population represents a unique neural network architecture. The genetic algorithm iteratively selects individuals from the population based on their fitness scores. [\ldots]'' & \textbf{The EvoGraph Architectural Foundry.} ``The EvoGraph Architectural Foundry is a specialized neuroevolution framework that forces a population of neural networks to solve the same problem using completely different structural logic. By utilizing a ``Novelty + Fitness'' selection criteria, it prevents the genetic algorithm from converging on a single ``best'' model, instead cultivating a diverse library of highly fit, structurally [\ldots]'' \\
\addlinespace[3pt]
Prompted naive$^\dagger$ & ``Particle Swarm Optimization: Use a particle swarm optimization algorithm to optimize the parameters of the neural networks. The algorithm starts with a population of particles, each representing a potential solution to the optimization problem. [\ldots]'' & \textbf{Behavioral Repulsion Particle Swarm Optimization (BR-PSO).} ``BR-PSO subverts this by introducing a Phenotypic Repulsion Force. Particles are still pulled toward high fitness, but they actively repel each other if they begin to solve the problem in the exact same way. This creates a ``Quality-Diversity'' swarm that scatters across the solution space, settling into multiple different local optima (diverse strategies) rather than [\ldots]'' \\
\bottomrule
\end{tabular}
\end{minipage}
\par\medskip

\par\medskip\noindent
\begin{minipage}{\textwidth}
\centering
\footnotesize
\setlength{\tabcolsep}{3pt}
\renewcommand{\arraystretch}{1.0}
\captionof{table}{7B vaccines: example seeds and their frontier-model expansions (verbatim; [\ldots] marks a cut). $^\dagger$Frontier model noted the seed had no concrete invention.}
\label{tab:examples-7b-vaccines}
\begin{tabular}{>{\raggedright\arraybackslash}p{0.10\textwidth}>{\raggedright\arraybackslash}p{0.42\textwidth}>{\raggedright\arraybackslash}p{0.42\textwidth}}
\toprule
\textbf{Model} & \textbf{Seed Idea} & \textbf{Refinement} \\
\midrule
Naive & ``A genetically modified strain of E. coli bacteria that produces a protein which, when ingested, boosts the immune system's ability to fight off infections by 40\%.'' & \textbf{The ``EnteroGuard'' Oral Immunization Vector (Strain EC-N1917-IG).} ``EnteroGuard is a live biotherapeutic product utilizing a genetically modified strain of Escherichia coli (specifically, the widely used and safe probiotic strain E. coli Nissle 1917). When ingested, this engineered bacteria transiently colonizes the gastrointestinal tract and secretes a specialized fusion protein that stimulates the Gut-Associated Lymphoid Tissue (GALT), effectively boosting the body's innate immunity [\ldots]'' \\
\addlinespace[3pt]
Naive & ``A pill that, when taken daily, builds up a tolerance to pathogens, preventing them from causing illness.'' & \textbf{Micro-Titer Enteric Immunization Pill (MTEIP).} ``The MTEIP reimagines vaccination as a 30-day daily pill regimen. By micro-dosing the vaccine over a month, the body is subjected to a gradual, sustained presentation of antigens. This mimics natural, low-level environmental exposure, building robust immunity with near-zero acute side effects.'' \\
\addlinespace[3pt]
Naive & ``A device that emits an electromagnetic field around it, destroying any biological organisms - including viruses and bacteria - within its range.'' & \textbf{The Tuned Electromagnetic Pathogen Inactivation System (TEPIS).} ``The TEPIS adapts the seed's concept of an ``organism-destroying electromagnetic field'' into a highly controlled, micro-scale manufacturing device. Instead of a macroscopic forcefield, TEPIS generates a precisely tuned, high-intensity electromagnetic (EM) field inside a containment tube. [\ldots]'' \\
\addlinespace[3pt]
Baseline$^\dagger$ & ``Get vaccinated: Vaccines can help protect against infectious diseases by stimulating the body's immune response.'' & \textbf{Optimized Immunization and Pharmacovigilance Protocol.} ``To maximize the advice to ``get vaccinated,'' this proposal outlines a structured, biologically detailed immunization protocol. It focuses on how a vaccine physically stimulates the immune response (efficacy) and the logistical framework required to monitor the patient post-administration (safety).'' \\
\addlinespace[3pt]
Prompted naive$^\dagger$ & ``Create a vector control program that aims to reduce the transmission of the infectious disease by targeting and controlling the populations of the disease vectors.'' & \textbf{Vector-Targeted Environmental Immunization (VTEI) System.} ``The core of the VTEI system is a specialized ``entomological vaccine.'' It utilizes a recombinant, insect-specific virus---such as a genetically modified Densovirus---which is entirely harmless to humans and vertebrates but highly infectious to the target mosquito species.'' \\
\bottomrule
\end{tabular}
\end{minipage}
\par\medskip

\par\medskip\noindent
\begin{minipage}{\textwidth}
\centering
\footnotesize
\setlength{\tabcolsep}{3pt}
\renewcommand{\arraystretch}{1.0}
\captionof{table}{7B backpropagation: example seeds and their frontier-model expansions (verbatim; [\ldots] marks a cut). $^\dagger$Frontier model noted the seed had no concrete invention.}
\label{tab:examples-7b-backpropagation}
\begin{tabular}{>{\raggedright\arraybackslash}p{0.10\textwidth}>{\raggedright\arraybackslash}p{0.42\textwidth}>{\raggedright\arraybackslash}p{0.42\textwidth}}
\toprule
\textbf{Model} & \textbf{Seed Idea} & \textbf{Refinement} \\
\midrule
Naive$^\dagger$ & ``Regularize the network: One way to prevent overfitting and ensure that the neural network can generalize well to new, unseen data is to regularize the network. This could involve adding a regularization term to the loss function, which penalizes large weights in the network, or applying techniques such as dropout to randomly deactivate a fraction [\ldots]'' & \textbf{Stochastic Gradient Masking (Dropout) Engine for Backpropagation.} ``Based on the seed's premise of ``randomly deactivating a fraction of the neurons during training,'' this proposal details the engineering mechanism of a Stochastic Gradient Masking Engine (commonly known as Dropout). The system forces a neural network to learn robust, independent features by ensuring that no neuron can rely on the presence of another specific [\ldots]'' \\
\addlinespace[3pt]
Baseline$^\dagger$ & ``Curriculum Learning: Design a curriculum that gradually increases the complexity of the images. Start with simple shapes and patterns, and gradually introduce more complex images.'' & \textbf{Dynamic Morphological Curriculum Training (DMCT).} ``DMCT refines the seed's idea of ``starting with simple shapes and patterns'' by automating this process directly within the backpropagation loop. Instead of changing the dataset, DMCT uses a differentiable image-processing layer that dynamically strips complex images down to basic geometric shapes, gradually reintroducing complexity (textures, high-frequency details, and colors) strictly based on the stabilization [\ldots]'' \\
\addlinespace[3pt]
Prompted naive$^\dagger$ & ``Ensemble Learning: Train multiple independent neural networks using different training data subsets or different initialization parameters. The individual networks learn to recognize different patterns and features within the input data. The final output is obtained by combining the predictions from all the individual networks, typically using a weighted average or majority voting approach.'' & \textbf{Orthogonal Gradient Routing for Ensemble Specialization (OGRES).} ``The seed notes that in an ensemble, ``individual networks learn to recognize different patterns and features.'' In standard machine learning, this specialization is a passive byproduct of giving models ``different training data subsets or different initialization parameters.'' OGRES (Orthogonal Gradient Routing for Ensemble Specialization) refines this into an active, unified backpropagation framework. [\ldots]'' \\
\bottomrule
\end{tabular}
\end{minipage}
\par\medskip

\section{Additional Per-Prompt Embeddings} \label{app:add_embed}

\begin{figure}[h]
    \centering
    \begin{subfigure}[ht]{0.48\textwidth}
        \centering
        \includegraphics[width=1.0\linewidth]{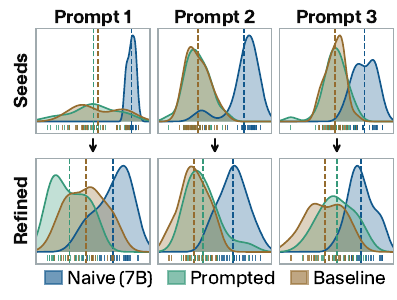}
        \caption{Dental Hygiene Embeddings}
        \label{fig:7b_dental_embed}
    \end{subfigure}
    \begin{subfigure}[ht]{0.48\textwidth}
        \centering
        \includegraphics[width=1.0\linewidth]{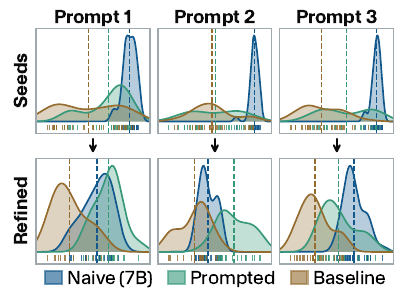}
        \caption{Neuroevolution Embeddings}
        \label{fig:7b_neuro_embed}
    \end{subfigure}
    \begin{subfigure}[ht]{0.48\textwidth}
        \centering
        \includegraphics[width=1.0\linewidth]{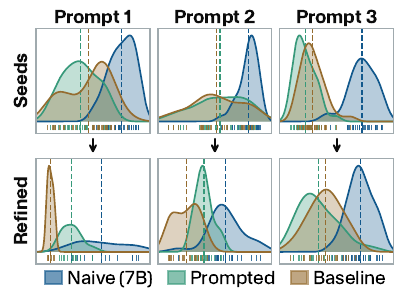}
        \caption{Vaccination Embeddings}
        \label{fig:7b_vaccines_embed}
    \end{subfigure}
    \caption{\textbf{Additional 7B model embeddings.}  Same format as Figure \ref{fig:1b_dental_embeddings}.  These show per-prompt embeddings not shown in the main paper.  They continue the trend of the naive model generally clustering away from the baselines.}
    \label{fig:7b_model_embeddings}
\end{figure}

Additional response embeddings for the 7B models not shown in Figure \ref{fig:7b_embedding} are shown in Figure \ref{fig:7b_model_embeddings}.

\end{document}